\documentclass{article} % For LaTeX2e
\usepackage{iclr2023_conference,times}

\usepackage{amsmath,amsfonts,bm}

\def\eqref#1{equation~\ref{#1}}
\def\1{\bm{1}}

\DeclareMathAlphabet{\mathsfit}{\encodingdefault}{\sfdefault}{m}{sl}
\SetMathAlphabet{\mathsfit}{bold}{\encodingdefault}{\sfdefault}{bx}{n}

\usepackage{hyperref}
\usepackage{url}

\usepackage[utf8]{inputenc} % allow utf-8 input
\usepackage[T1]{fontenc}    % use 8-bit T1 fonts
\usepackage{booktabs}       % professional-quality tables
\usepackage{amsfonts}       % blackboard math symbols
\usepackage{nicefrac}       % compact symbols for 1/2, etc.
\usepackage{microtype}      % microtypography
\usepackage{xcolor}         % colors
\usepackage{multirow} 
\usepackage{bbding}
\usepackage{subfigure}
\usepackage{graphicx}
\usepackage{wrapfig}
\usepackage{enumitem}

 \iclrfinalcopy
\author{
  Rongjie Huang$^{1}$\thanks{Equal Contribution}, ~~Jinglin Liu$^{1}$\footnotemark[1], ~~Huadai Liu$^{1}$\footnotemark[1], ~~Yi Ren$^{2}$, ~~Lichao Zhang$^{1}$,  \\ \textbf{~Jinzheng He$^{1}$, ~~Zhou Zhao$^{1}$\thanks{Corresponding Author}} \\ 
  \\
  $^{1}$Zhejiang University\\
  \texttt{\{rongjiehuang, jinglinliu, huadailiu, zhaozhou\}@zju.edu.cn}\\
  \\
$^{2}$ByteDance\\
\texttt{ren.yi@bytedance.com}\\
}

\title{TranSpeech: Speech-to-Speech Translation With Bilateral Perturbation}

\begin{document}

\maketitle

\begin{abstract}
  Direct speech-to-speech translation (S2ST) with discrete units leverages recent progress in speech representation learning. Specifically, a sequence of discrete representations derived in a self-supervised manner are predicted from the model and passed to a vocoder for speech reconstruction, while still facing the following challenges: 1) Acoustic multimodality: the discrete units derived from speech with same content could be indeterministic due to the acoustic property (e.g., rhythm, pitch, and energy), which causes deterioration of translation accuracy; 2) high latency: current S2ST systems utilize autoregressive models which predict each unit conditioned on the sequence previously generated, failing to take full advantage of parallelism. In this work, we propose TranSpeech, a speech-to-speech translation model with bilateral perturbation. To alleviate the acoustic multimodal problem, we propose bilateral perturbation (BiP), which consists of the style normalization and information enhancement stages, to learn only the linguistic information from speech samples and generate more deterministic representations. With reduced multimodality, we step forward and become the first to establish a non-autoregressive S2ST technique, which repeatedly masks and predicts unit choices and produces high-accuracy results in just a few cycles. Experimental results on three language pairs demonstrate that BiP yields an improvement of 2.9 BLEU on average compared with a baseline textless S2ST model. Moreover, our parallel decoding shows a significant reduction of inference latency, enabling speedup up to 21.4x than autoregressive technique. \footnote{Audio samples are available at \url{https://TranSpeech.github.io/}.}

\end{abstract}

\section{Introduction}
% which are purely spoken and have no written text
Speech-to-speech translation (S2ST) aims at converting speech from one language into speech in another, significantly breaking down communication barriers between people not sharing a common language. Among the conventional method~\citep{lavie1997janus,nakamura2006atr,wahlster2013verbmobil}, the cascaded system of automatic speech recognition (ASR), machine translation (MT), or speech-to-text translation (S2T) followed by text-to-speech synthesis (TTS) have demonstrated reasonable results yet suffering from expensive computational costs. Compared to these cascaded systems, recently proposed direct S2ST literature~\citep{jia2019direct,zhang2020uwspeech,jia2021translatotron,lee2021direct,lee2021textless} demonstrate the benefits of lower latencies as fewer decoding stages are needed.

% further supporting translation between unwritten languages. 

% In textless speech-to-speech translation, our goal is mainly two-fold:
% \begin{itemize}
%     \item High-accuracy: achieving strong translation performance without the use of any text or phoneme data is a challenging problem especially when considering the multimodality issue, where the speech with same content could be different due to a variety of acoustic condition. It is vital for S2ST models to capture the linguistic content information for accurate translation. 
%     \item Fast: low latency is essential when considering real-time application. This poses a challenge for all neural speech-to-speech translators.
% \end{itemize}

\begin{figure*}
    \centering
    \vspace{-6mm}
    \includegraphics[width=0.98\textwidth,trim={1.0cm 0cm 1.0cm 0cm}]{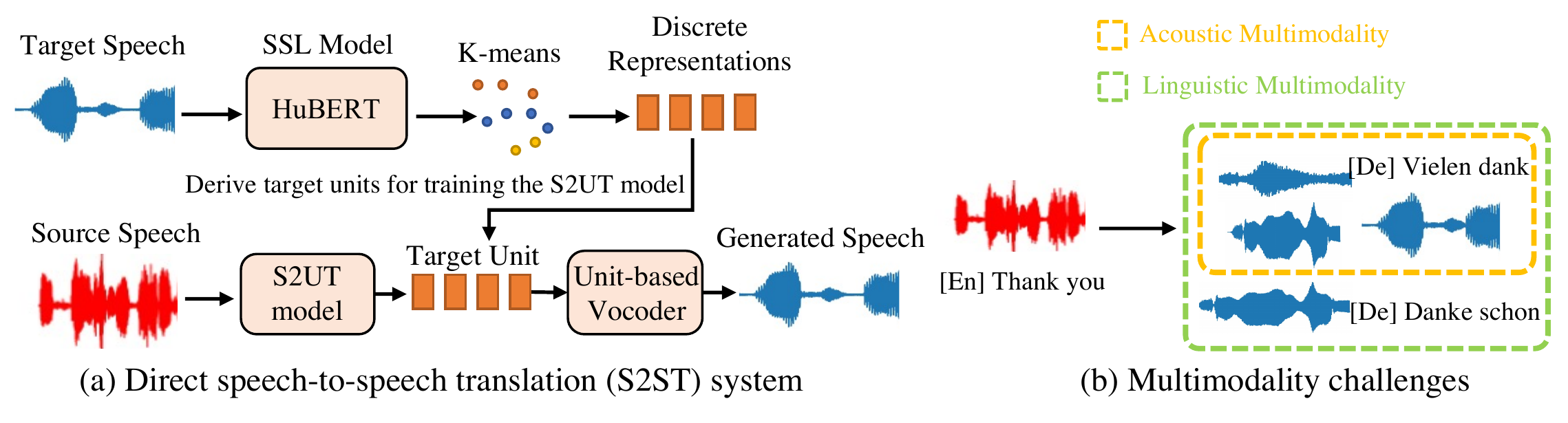}
    \vspace{-3mm}
   \caption{1) Acoustic multimodality: Speech with the same content "Vielen dank" could be different due to a variety of acoustic conditions; 2) Linguistic multimodality~\citep{gu2017non,wang2019non}: There are multiple correct target translations ("Danke schon" and "Vielen dank") for the same source word/phrase/sentence ("Thank you"). } 
   % \vspace{-4mm}
    \label{fig:pipeline}
  \end{figure*}

Among them, ~\citet{lee2021direct,lee2021textless} leverage recent progress on self-supervised discrete units learned from unlabeled speech for building textless S2ST systems, further supporting translation between unwritten languages. As illustrated in Figure~\ref{fig:pipeline}(a), the unit-based textless S2ST system consists of a speech-to-unit translation (S2UT) model followed by a unit-based vocoder that converts discrete units to speech, leading to a significant improvement over previous literature. 

In modern textless speech-to-speech translation (S2ST), our goal is mainly two-fold: 1) high quality: direct S2ST is challenging, especially without using the transcription. 2) low latency: high inference speed is essential when considering real-time translation. However, the current development of the unit-based textless S2ST system is hampered by two major challenges: 1) It is challenging to achieve high translation accuracy due to the acoustic multimodality (as illustrated in the orange dotted box in Figure~\ref{fig:pipeline}(b)): different from the language tokens (e.g., bpe) used in the text translation, the self-supervised representation derived from speech with the same content could be different due to a variety of acoustic conditions (e.g., speaker identity, rhythm, pitch, and energy), including both linguistic content and acoustic information. As such, the indeterministic training target for speech-to-unit translation fails to yield good results; and 2) Building a parallel model upon multimodal S2ST systems with reasonable accuracy is challenging as it introduces further indeterminacy. A non-autoregressive (NAR) S2ST system generates all tokens in parallel without any limitation of sequential dependency, making it a poor approximation to the actual target distribution. With the acoustic multimodality unsettled, the parallel decoding approaches increasingly burden S2ST capturing the distribution of target translation.

% As illustrated in the orange dotted box in Figure~\ref{fig:pipeline}(c), this multimodal problem poses a challenge for accurate translation. 
% speech encoder pre-trained on unlabeled speech to convert target speech into discrete units, where a sequence of discrete units is directly generated from the translation model. 
% , where ~\citep{gu2017non,wang2019non}. 

% Therefore, this acoustic multimodality poses a challenge for generating deterministic representations for accurate translation.
% As such, this linguistic multimodality~\citep{gu2017non,wang2019non} prevents models from properly capturing the distribution of target 
  
% \begin{itemize}
%     \item Different from the language tokens (e.g., bpe) used in text translation, the self-supervised representation derived from speech with the same content could be different due to a variety of acoustic conditions (e.g., speaker identity, rhythm, pitch, and energy), including both linguistic content and acoustic information. As illustrated in the orange dotted box in Figure~\ref{fig:pipeline}(c), this multimodal problem poses a challenge for accurate translation. 
%     \item Current state-of-the-art S2ST utilizes the left-to-right autoregressive model, which predicts each unit conditioned on the sequence previously generated and be criticized for the low latency. The acoustic and linguistic multimodality (See Figure~\ref{fig:pipeline}(c)) further hinders the application of non-autoregressive sequence generation in S2ST. 
% \end{itemize}
% \vspace{-2mm}
% , enabling the translator to make accurate prediction multiple correct target translations for the same source word/phrase/sentence, also known as 
In this work, we propose TranSpeech, a fast speech-to-speech translation model with bilateral perturbation. To tackle the acoustic multimodal challenge, we propose a \textbf{Bilateral Perturbation (BiP)} technique that finetunes a self-supervised speech representation learning model with CTC loss to generate deterministic representation agnostic to acoustic variation. Based on preliminary speech analysis by decomposing a signal into linguistic and acoustic information, the bilateral perturbation consists of the 1) \textbf{style normalization} stage, which eliminates the acoustic-style information in speech and creates the style-agnostic "pseudo text" for finetuning; and 2) \textbf{information enhancement} stage, which applies information bottleneck to create speech samples variant in acoustic conditions (i.e., rhythm, pitch, and energy) while preserving linguistic information. The proposed bilateral perturbation guarantees the speech encoder to learn only the linguistic information from acoustic-variant speech samples, significantly reducing the acoustic multimodality in unit-based S2ST.
%  and enables the translator to make an accurate prediction. and guarantees the speech encoder to learn only the linguistic information in acoustic-variant speech samples. 
% As such, few models have been proposed to take full advantage of parallelism during inference in direct speech-to-speech translation. To accelerate the inference procedure, d

% , while real-time interpretation poses a challenge for all S2ST systems

The proposed bilateral perturbation eases acoustic multimodality and makes it possible for NAR generation. As such, we further step forward and become the first to establish a NAR S2ST technique, which repeatedly masks and predicts unit choices and produces high-accuracy results in just a few cycles. Experimental results on three language pairs demonstrate that BiP yields an improvement of 2.9 BLEU on average compared with baseline textless S2ST models. The parallel decoding algorithm requires as few as 2 iterations to generate samples that outperformed competing systems, enabling a speedup by up to 21.4x compared to the autoregressive baseline. TranSpeech further enjoys a speed-performance trade-off with advanced decoding choices, including multiple iterations, length beam, and noisy parallel decoding, trading by up to 3 BLEU points in translation results. The main contributions of this work include:

\begin{itemize} [leftmargin=*]
    \item Through preliminary speech analysis, we propose bilateral perturbation which assists in generating deterministic representations agnostic to acoustic variation. This novel technique alleviates the acoustic multimodal challenge and leads to significant improvement in S2ST.
    \item We step forward and become the first to establish a non-autoregressive S2ST technique with a mask-predict algorithm to speed up the inference procedure. To further reduce the linguistic multimodality in NAR translation, we apply the knowledge distillation technique and construct a less noisy and more deterministic corpus.
    % To the best of our knowledge, we are the first to establish a non-autoregressive S2ST technique that can translate samples in parallel.
    \item Experimental results on three language pairs demonstrate that BiP yields the promotion of 2.9 BLEU on average compared with baseline textless S2ST models. In terms of inference speed, our parallel decoding enables speedup up to 21.4x compared to the autoregressive baseline. 
    % with translation accuracy gain 
\end{itemize}

\section{Background: Direct Speech-to-Speech Translation}

Direct speech-to-speech translation has made huge progress to date. Translatotron~\citep{jia2019direct} is the first direct S2ST model and shows reasonable translation accuracy and speech naturalness. Translatotron 2~\citep{jia2021translatotron} utilizes the auxiliary target phoneme decoder to promote translation quality but still needs phoneme data during training. UWSpeech~\citep{zhang2020uwspeech} builds the VQ-VAE model and discards transcript in the target language, while paired speech and phoneme corpora of written language are required. 

Most recently, a direct S2ST system~\citep{lee2021direct} takes advantage of self-supervised learning (SSL) and demonstrates the results without using text data. However, the majority of SSL models are trained by reconstructing~\citep{chorowski2019unsupervised} or predicting unseen speech signals~\citep{chung2019unsupervised}, which would inevitably include factors unrelated to the linguistic content (i.e., acoustic condition). As such, the indeterministic training target for speech-to-unit translation fails to yield good results. The textless S2ST system~\citep{lee2021textless} further demonstrates to obtain the speaker-invariant representation by finetuning the SSL model to disentangle the speaker-dependent information. However, this system only constrains speaker identity, and the remaining aspects (i.e., content, rhythm, pitch, and energy) are still lumped together. 

At the same time, various approaches that perturb information flow to fine-tune acoustic models have demonstrated efficiency in promoting downstream performance. A line of works~\citep{yang2021voice2series,gao2022wavprompt} utilizes the pre-trained encoder and introduces approaches that reprogram acoustic models in downstream tasks. For multi-lingual tuning, ~\citet{yen2021study} propose a novel adversarial reprogramming approach for low-resource spoken command recognition (SCR). Sharing a common insight, we tune a pre-trained acoustic model with bilateral perturbation technique and generates more deterministic units agnostic to acoustic conditions, including rhythm, pitch, and energy. Following the common textless setup in Figure~\ref{fig:pipeline}(a), we design a challenging NAR S2ST technique especially for applications requiring low latency. More details have been attached in Appendix~\ref{related}.
\section{Speech Analysis and Bilateral Perturbation}

% The majority of SSL models are trained by reconstructing~\citep{chorowski2019unsupervised} or predicting unseen speech signals~\citep{chung2019unsupervised}, which would inevitably include factors unrelated to the linguistic content (i.e., acoustic condition). 

\subsection{Acoustic Multimodality}
As reported in previous textless S2ST system~\citep{lee2021textless}, speech representations predicted by the self-supervised pre-trained model include both linguistic and acoustic information. As such, derived representations of speech samples with the same content can be different due to the acoustic variation, and the indeterministic training target for speech-to-unit translation (as illustrated in Figure~\ref{fig:pipeline}(a)) fails to yield good results. To address this multimodal issue, we conduct a preliminary speech analysis and introduce the bilateral perturbation technique. More details on how indeterminacy units influence S2ST have been attached in Appendix~\ref{indeterministic}.
% fail to yield good results in the translation stage and 

\subsection{Speech Analysis}

In this part, we decompose speech variations~\citep{cui2022varietysound,huang2021multi,yang2022norespeech} into linguistic content and acoustic condition (e.g., speaker identity, rhythm, pitch, and energy) and provide a brief primer on each of these components.

\textbf{Linguistic Content} represents the meaning of speech signals. To translate a speech sample to another language, learning the linguistic information from the speech signal is crucial; \textbf{Speaker Identity} is perceived as the voice characteristics of a speaker. 
% It is reflected by the formant frequencies, which are the resonant frequency components in the vocal tract; 
\textbf{Rhythm} characterizes how fast the speaker utters each syllable, and the duration plays a vital role in acoustic variation; \textbf{Pitch} is an essential component of intonation, which is the result of a constant attempt to hit the pitch targets of each syllable; \textbf{Energy} affects the volume of speech, where stress and tone represent different energy values. 
% indicates the frame-level magnitude of mel-spectrograms and 
% In practice, we regard the L2-norm of the amplitude of each short-time Fourier transform (STFT) frame as energy.

% $S, \overline{S}, \hat{S}$ to denote the original speech and perturbed speeches used in style normalization and information enhancement, respectively. 
\begin{figure}[]
    \centering
    \small
    \vspace{-10mm}
    \includegraphics[width=0.9\textwidth,trim={1.0cm 0cm 1.0cm 0cm}]{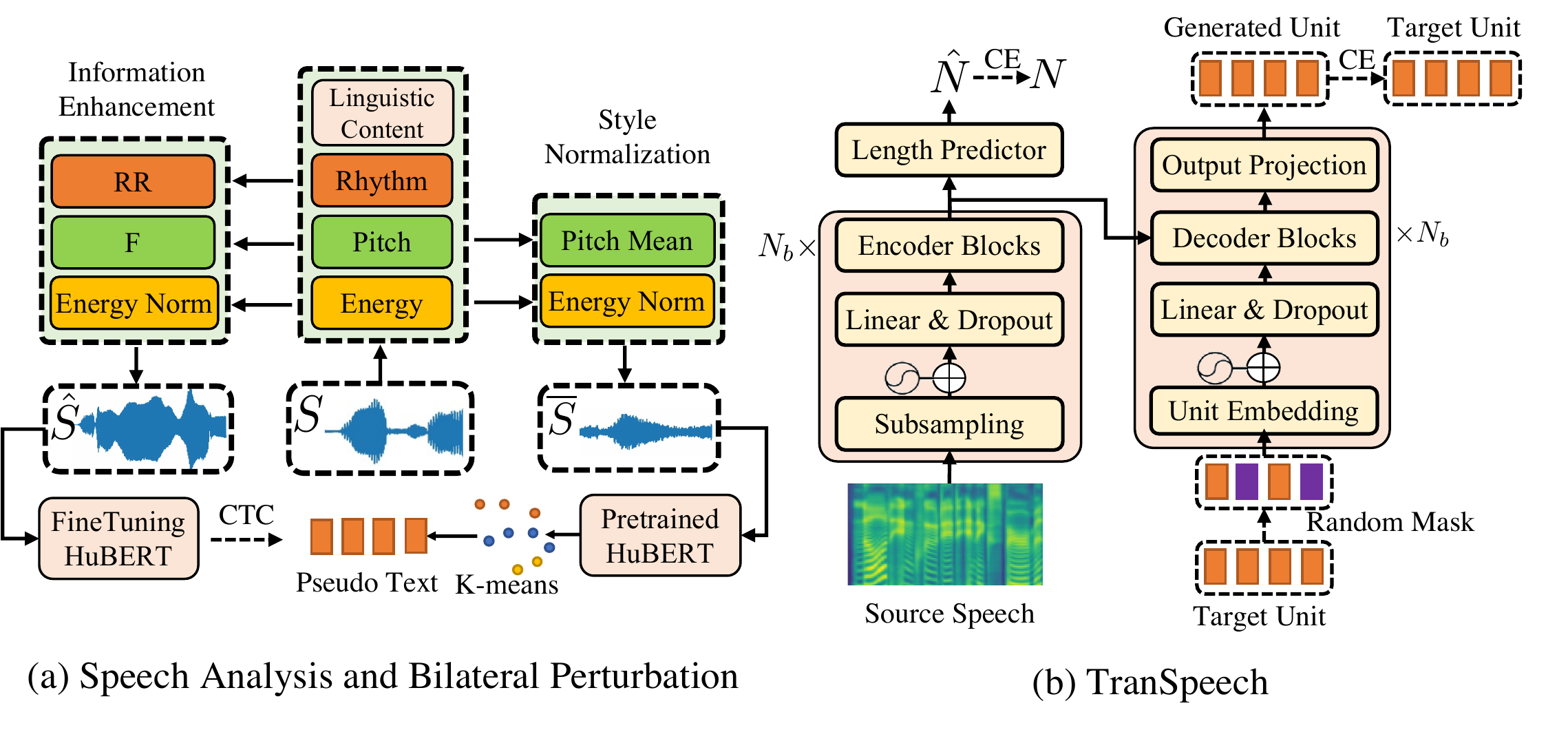}
    \vspace{-5mm}
    \caption{In subfigure(a), we use $RR$ and $F$ to respectively denote the random resampling and a chain function for random pitch shifting. In subfigure(b), the "sinusoidal-like symbol" denotes the positional encoding, we have $N_b$ encoder and decoder blocks. During training, we randomly select the masked position and compute the cross-entropy loss (denoted as "CE").} 
    % \vspace{-4mm}
    \label{fig:norm}
\end{figure}

\subsection{Bilateral Perturbation}

% To alleviate the multimodality problem and promote the translation accuracy of textless S2ST system, we propose a bilateral perturbation technique to generate more deterministic training targets which are agnostic to acoustic variation for speech-to-unit translation. 
To alleviate the multimodal problem and increase the translation accuracy in the S2ST system, we propose bilateral perturbation that disentangles the acoustic variation and generates deterministic speech representations according to the linguistic content. Specifically, we leverage the success of connectionist temporal classification (CTC) finetuning~\citep{baevski2019effectiveness} with a pre-trained speech encoder, using the perturbed input speech and normalized target. Since how to obtain speaker-invariant representation has been well-studied~\citep{lee2021textless,hsu2020text}, we focus on the more challenging acoustic conditions in a single-speaker scenario, including rhythm, pitch, and energy variations.

\subsubsection{Overview}
% the discrete unit-based bilateral perturbation
Denote the domain of speech samples by $S \subset \mathbb{R}$ and the perturbed speeches in style normalization and information enhancement by $\overline{S}, \hat{S}$ respectively. The source language is therefore a sequence of speech samples $X=\left\{x_{1}, \ldots, x_{N^{\prime}}\right\}$, where $N^{\prime}$ is the number of frames in source speech. The SSL model is composed of a multi-layer convolutional feature encoder $f$ which takes as input raw audio $S$ and outputs discrete latent speech representations. In the end, the audio in the target language is represented as discrete units $Y=\{y_{1}, \ldots, y_{N}\}$, where $N$ is the number of units. The overview of the information flow is shown in Figure~\ref{fig:norm}(a), and we consider tackling the multimodality in bilateral sides for CTC finetuning, including 1) \textbf{style normalization} stage to eliminate the acoustic information in the CTC target and create the acoustic-agnostic "pseudo text"; and 2) \textbf{information enhancement} stage which applies bottleneck on acoustic features to create speech samples variant in acoustic conditions (e.g., rhythm, pitch, and energy) while preserving linguistic content information. In the final, we train an ASR model using the perturbed speech $\hat{S}$ as input and the "pseudo text" as the target. 
% As such, according to speeches with acoustic variation, the speech encoder is encouraged to learn the "average" information referring to deterministic linguistic content.
As a result, according to speeches with acoustic variation, the ASR model with CTC decoding is encouraged to learn the "average" information referring to linguistic content and generate deterministic representations, significantly reducing multimodality and promoting speech-to-unit translation. In the following subsections, we present the bilateral perturbation technique in detail:
% The proposed bilateral perturbation guarantees the speech encoder to learn only the linguistic information from acoustic-variant speech samples, significantly reducing the acoustic multimodality in unit-based S2ST. 
% could learn only the linguistic information in speech samples and generate deterministic representations, significantly reducing multimodality and promoting speech-to-unit translation. We describe the bilateral perturbation techniques in detail:
% \vspace{-3mm}
\subsubsection{Style Normalization}
To create the acoustic-agnostic "pseudo text" for CTC finetuning, the acoustic-style information should be eliminated and disentangled: 1) We first compute the averaged pitch fundamental frequency $\overline{p}$ and energy $\overline{e}$ values in original dataset $S$; and 2) for each sample in $S$, we conduct pitch shifting to $\overline{p}$ and normalize its energy to $\overline{e}$, resulting in a new dataset $\overline{S}$ with the averaged acoustic condition, where the style-specific information has been eliminated; finally, 3) the self-supervised learning (SSL) model encodes $\overline{S}$ and creates the normalized targets for CTC finetuning.

\subsubsection{Information Enhancement} \label{info_perturbation}
% We would like to train the ASR model to selectively extract only the linguistic information from the input waveform, not acoustic-related information. 
According to the speech samples with different acoustic conditions, the ASR model is supposed to learn the deterministic representation referring to linguistic content. As such, we apply the following functions as information bottleneck on acoustic features (e.g., rhythm, pitch, and energy) to create highly acoustic-variant speech samples $\hat{S}$, while the linguistic content remains unchanged, including 1) formant shifting $fs$, 2) pitch randomization $pr$, 3) random frequency shaping using a parametric equalizer $peq$, and 4) random resampling $RR$. 
\begin{itemize} [leftmargin=*]
    \item For rhythm information, random resampling $RR$ divides the input into segments of random lengths, and we randomly stretch or squeeze each segment along the time dimension. 
    \item For pitch information, we apply the chain function $F = fs(pr(peq(S)))$ to randomly shift the pitch value of original speech $S$. 
    \item For energy information, we perturb the audio in the waveform domain.
\end{itemize}

% with information bottleneck 
The perturbed waveforms $\hat{S}$ are highly variant on acoustic features (i.e., rhythm, pitch, and energy) while preserving linguistic information. It guarantees the speech encoder to learn the "acoustic-averaged" information referring to linguistic content and generate deterministic representations. The hyperparameters of the perturbation functions have been included in Appendix~\ref{perturbation}.
% In the end, t
% according to a variety of input speeches with different acoustic condition
% As such, we apply information bottleneck to perturb the acoustic information included in input waveform $S$, which guarantees the speech encoder to generate more deterministic "average" representations according to a variety of input speeches with different acoustic condition. 
% \vspace{-3mm}

\section{TranSpeech}

% As illustrated in Figure~\ref{fig:norm}(b), we present the key features in TranSpeech, including the conformer encoder and non-autoregressive unit decoder with mask-predict algorithm. 
The S2ST pipeline has been illustrated in Figure~\ref{fig:norm}(a), we 1) use the SSL HuBERT~\citep{hsu2021hubert} tuned by BiP to derive discrete units of target speech; 2) build the sequence-to-sequence model TranSpeech for speech-to-unit translation (S2UT) and 3) apply a separately trained unit-based vocoder to convert the translated units into waveform.

In this section, we first overview the encoder-decoder architecture for TranSpeech, following which we introduce the knowledge distillation procedure to alleviate the linguistic multimodal challenges. Finally, we present the mask-predict algorithm in both training and decoding procedures and include more advanced decoding choices.
% The non-autoregressive model with mask-predict algorithm repeatedly reconsiders unit choices and produces high-quality translations in just a few cycles, taking full advantage of parallelism in inference.

\subsection{Architecture}

The overall architecture has been illustrated in Figure~\ref{fig:norm}(b), and we put more details on the encoder and decoder block in Appendix~\ref{appendix:arch}.

\textbf{Conformer Encoder.}
Different from previous textless S2ST literature~\citep{lee2021textless}, we use conformer blocks~\citep{gulati2020conformer} in place of transformer blocks~\citep{vaswani2017attention}. The conformer model~\citep{guo2021recent,chen2021continuous} has demonstrated its efficiency in combining convolution neural networks and transformers to model both local and global dependencies of audio in a parameter-efficient way, achieving state-of-the-art results on various downstream tasks. Furthermore, we employ the multi-head self-attention with a relative sinusoidal positional encoding scheme from Transformer-XL~\citep{dai2019transformer}, which promotes the robustness of the self-attention module and generalizes better to different utterance lengths. 

% A conformer block is composed of four modules stacked together, i.e, a feed-forward module, a self-attention module, a convolution module, and a second feed-forward module in the end. We employ multi-headed self-attention with relative sinusoidal positional encoding scheme from Transformer-XL~\citep{dai2019transformer}, which promotes the robustness of the self-attention module and generalizes better on different utterance length. 

\textbf{Non-autoregressive Unit Decoder.}
% target units $Y_{mask}$ given the encoder output $X$ and part of the target units $Y_{obs}$ in parallel. In non-autoregressive decoding, it makes strong assumption that the target units are conditionally independent of each other, and the individual probabilities $P(y|X, Y_{obs})$ are predicted for each $y \in Y_{mask}$. Since the number of units in $Y_{mask}$ should be given in advance, TranSpeech further predicts the length of the target sequence $N$ by pooling the encoder outputs into a length classifier~\citep{lee2018deterministic}.
Currently, S2ST systems utilize the autoregressive S2UT models and suffer from high inference latency. Given the $N^{\prime}$ frames source speech $X=\left\{x_{1}, \ldots, x_{N^{\prime}}\right\}$, autoregressive model $\theta$ factors the distribution over possible outputs $Y=\left\{y_{1}, \ldots, y_{N}\right\}$ by $p(Y \mid X ; \theta)=\prod_{i=1}^{N+1} p(y_{i} \mid y_{0: i-1}, x_{1: N^{\prime}} ; \theta),$
% \begin{shrinkeq}{-3ex}
% \vspace{-2mm}
% \begin{equation}
%     \small
%     p(Y \mid X ; \theta)=\prod_{i=1}^{N+1} p(y_{i} \mid y_{0: i-1}, x_{1: N^{\prime}} ; \theta),
% \end{equation} 
% \end{shrinkeq}
where the special tokens $y_{0} (\langle bos \rangle)$ and $y_{N+1} (\langle eos \rangle)$ are used to represent the beginning and end of all target units. 

Unlike the relatively well-studied non-autoregressive (NAR) MT~\citep{gu2017non,wang2019non,gu2019levenshtein,ghazvininejad2019mask,GASLT}, building NAR S2UT models that generate units in parallel could be much more challenging due to the joint linguistic and acoustic multimodality. Yet the proposed bilateral perturbation eases this acoustic multimodality and makes it possible for NAR modeling. As such, we further step forward and become the first to establish a NAR S2ST model $\theta$. 

It assumes that the target sequence length $N$ can be modeled with a separate conditional distribution $p_{L}$, and the distribution becomes $p(Y \mid X ; \theta)=p_{L}(T \mid x_{1: N^{\prime}} ; \theta) \cdot \prod_{i=1}^{N} p\left(y_{i} \mid x_{1: N^{\prime}} ; \theta\right)$.
% \begin{shrinkeq}{-3ex}
%     \vspace{-2mm}
% \begin{equation}
%     \small
%     p(Y \mid X ; \theta)=p_{L}(T \mid x_{1: N^{\prime}} ; \theta) \cdot \prod_{i=1}^{N} p\left(y_{i} \mid x_{1: N^{\prime}} ; \theta\right)
%     \end{equation}
% \end{shrinkeq}
The target units are conditionally independent of each other, and the individual probabilities $p$ is predicted for each token in $Y$. Since the length of target units $N$ should be given in advance, TranSpeech predicts it by pooling the encoder outputs into a length predictor.
% ~\citep{lee2018deterministic} as illustrated in Figure~\ref{fig:norm}(b)
% In non-autoregressive decoding, it makes a strong assumption that the target units are conditionally independent of each other, and the individual probabilities $p_i$ are predicted for each token $Y_i$. Since the number of target units should be given in advance, 
% TranSpeech further predicts the length of the target sequence $N$ by pooling the encoder outputs into a length predictor~\citep{lee2018deterministic}.
% predict each unit conditioned on the previously generated sequence $y$, the proposed unit decoder predicts a set of target units in parallel. 

\subsection{Linguistic Multimodality} \label{section:distill}
% \footnote{Previous NAR MT literature~\citep{gu2017non,wang2019non} alleviates the linguistic multimodality by constructing a less noisy and more deterministic sampled translation corpus via knowledge distillation.}

% NAR models generate all tokens in parallel without any limitation of sequential dependency, making it a poor approximation to the actual target distribution, which exhibits a strong correlation across time. 

As illustrated in Figure~\ref{fig:pipeline}(b), there might be multiple valid translations for the same source utterance, and thus this linguistic multimodality degrades the ability of NAR models to properly capture the target distribution. To alleviate this linguistic multimodality in NAR translation, we apply knowledge distillation to construct a sampled translation corpus from an autoregressive teacher, which is less noisy and more deterministic than the original one. The knowledge of the AR model is distilled to the NAR model, assisting to capture the target distribution for better accuracy.

\subsection{Mask-Predict}
The NAR unit decoder applies the mask-predict algorithm~\citep{ghazvininejad2019mask} to repeatedly reconsider unit choices and produce high-accuracy translation results in just a few cycles. 

\textbf{Training.}
% we randomly select $Y_{mask}$ among targets: 
During training, the target units are given conditioned on source speech sample $X$ and the unmasked target units $Y_{obs}$. As illustrated in Figure~\ref{fig:norm}(b), given the length $N$ of the target sequence, we first sample the number of masked units from a uniform distribution $n\sim\mathrm{Unif}(\{1,\cdots,N\})$, and then randomly choose the masked position. For the learning objective, we compute the cross-entropy (CE) loss with label smoothing between the generated and target units in masked places, and the CE loss for target length prediction is further added. 

\textbf{Decoding.}
In inference, the algorithm runs for pre-determined $T$ times of iterative refinement, and we perform a \textit{mask} operation at each iteration, followed by \textit{predict}.

In the first iteration $t=0$, we predict the length $N$ of target sequence and mask all units $Y=\left\{y_{1}, \ldots, y_{N}\right\}$. In the following iterations, we mask $n$ units with the lowest probability scores $p$:
\begin{equation}
    Y_{mask}^t =\arg \min _{i}(p_{i}, n) \qquad Y_{obs}^t = Y \backslash Y_{mask}^{t},
\end{equation}
where $n$ is a function of the iteration $t$, and we use linear decay $n=N \cdot \frac{T-t}{T}$ in this work.

After masking, TranSpeech predicts the masked units $Y_{mask}^t$ conditioned on the source speech $X$ and unmasked units $Y_{obs}$. We select the prediction with the highest probability $p$ for each $y_{i} \in Y_{mask}^t$ and update its probability score accordingly:
\begin{equation}
y_{i}^t=\arg \max_w  P(y_i=w \mid X, Y_{o b s}^t ; \theta) \qquad  p_{i}^t=\max_w P(y_i=w \mid X, Y_{o b s}^t ; \theta)
\end{equation}

\subsection{Advanced Decoding Choices}

\textbf{Target Length Beam.}
It has been reported~\citep{ghazvininejad2019mask} that translating multiple candidate sequences of different lengths can improve performance. As such, we select the top $K$ length candidates with the highest probabilities and decode the same example with varying lengths in parallel. In the following, we pick up the sequence with the highest average log probability as our result. It avoids distinctly increasing the decoding time since the computation can be batched.

\textbf{Noisy Parallel Decoding.}
The absence of the AR decoding procedure makes it more difficult to capture the target distribution in S2ST. To obtain the more accurate optimum of the target distribution and compute the best translation for each fertility sequence, we use the autoregressive teacher to identify the best overall translation. 

% which increases the computational resources required linearly by the sample size.

\section{Experiments}

\subsection{Experimental Setup} \label{setup}

Following the common practice in the direct S2ST pipeline, we apply the publicly-available pre-trained multilingual HuBERT (mHuBERT) model and unit-based HiFi-GAN vocoder~\citep{polyak2021speech,kong2020hifi} and leave them unchanged. 

% In this section, we present the experimental setup of TranSpeech with proposed bilateral perturbation.

% ~\citep{lee2021textless} pretrained on unlabeled speech as the basis for bilateral perturbation, and the pretrained unit-based HiFi-GAN vocoder~\citep{polyak2021speech,kong2020hifi} for unit-to-speech synthesis. 
% In this section, we focus on the experimental setup of the proposed bilateral perturbation and speech-to-unit translation model.

\textbf{Dataset.} 
For a fair comparison, we use the benchmark CVSS-C dataset~\citep{jia2022cvss}, which is derived from the CoVoST 2~\citep{wang2020covost} speech-to-text translation corpus by synthesizing the translation text into speech using a single-speaker TTS system. To evaluate the performance of the proposed model, we conduct experiments on three language pairs, including French-English (Fr-En), English-Spanish (En-Es), and English-French (En-Fr). 
% As describe in Section~\ref{}, we disentangling the rhythm, pitch and energy information in self-supervised representation in the single speaker scenario, and 
% Following previous work on sequence-level knowledge distillation~\citep{gu2017non,wang2019non}, we train the non-autoregressive student TranSpeech on translations produced by a standard left-to-right conformer teacher as described in Section~\ref{section:distill}. 

\textbf{Model Configurations and Training.}
For bilateral perturbation, we finetune the publicly-available mHuBERT model for each language separately with CTC loss until 25k updates using the Adam optimizer ($\beta_{1}=0.9, \beta_{2}=0.98, \epsilon=10^{-8}$). Following the practice in textless S2ST~\citep{lee2021textless}, we use the k-means algorithm to cluster the representation given by the well-tuned mHuBERT into a vocabulary of 1000 units. TranSpeech computes 80-dimensional mel-filterbank features at every 10-ms for the source speech as input, and we set $N_b$ to 6 in encoding and decoding blocks. In training the TranSpeech, we remove the auxiliary tasks for simplification and follow the unwritten language scenario. TranSpeech is trained until convergence for 200k steps using 1 Tesla V100 GPU. A comprehensive table of hyperparameters is available in Appendix~\ref{appendix:arch}. 

% trained with Adam optimizer ($\beta_{1}=0.9, \beta_{2}=0.98, \epsilon=10^{-8}$)
% The model with the best BLEU~\citep{papineni2002bleu} on the dev set is used for evaluation. 
% The conformer encoder uses a 16-layer conformer encoder with embedding size 512 and 8 attention heads. The unit decoder consists of 6 transformer layers with the same embedding size, attention heads and FFN embedding size as the encoder. 
% The k-means models are learned from the combination of En, Es and French speech data from VoxPopuli~\citep{wang2021voxpopuli}, which we empirically find to be universal cross dataset.

% In the end, the publicly-available universal unit-to-speech vocoder~\citep{polyak2021speech} enhanced with a duration prediction module recover the duration for the reduced unit sequence, and generate high-fidelity speech samples for each language separately.
\textbf{Evaluation and Baseline models.}
% Decoding is implemented in PyTorch on a single NVIDIA Tesla V100. generate discrete token sequences from source speech with 
% During inference, we use beam search with beam size to $5$ in autoregressive decoding. 
% As the ASR output is in lowercase and without digits and punctuation except apostrophes, we normalize both the reference text and the text output from the S2ST model before computing BLEU. 
For translation accuracy, we pre-train an ASR model to generate the corresponding text of the translated speech and then calculate the BLEU score~\citep{papineni2002bleu} between the generated and the reference text. In decoding speed, latency is computed as the time to decode the single n-frame speech sample averaged over the test set using 1 V100 GPU. 

We compare TranSpeech with other systems using the publicly-available \textit{fairseq} framework~\citep{ott2019fairseq}, including 1) Direct ASR, where we transcribe S2ST data with open-sourced ASR as reference and compute BELU; 2) Direct TTS, where we synthesize speech samples with target units, and then transcribe the speech to text and compute BELU; 3) S2T+TTS cascaded system, where we train the S2T basic transformer model~\citep{wang2020fairseqs2t} and then apply TTS model~\citep{ren2020fastspeech,kong2020hifi} for speech generation; 4) basic transformer~\citep{lee2021direct} without using text, and 5) basic norm transformer~\citep{lee2021textless} with speaker normalization.
% 3) S2T+TTS cascaded system, where we first train a S2T basic transformer model~\citep{wang2020fairseqs2t} for each language pair using the CoVoST 2 dataset without multitask learning, and then apply FastSpeech 2~\citep{ren2020fastspeech} with HiFi-GAN vocoder~\citep{kong2020hifi} for text-to-spectrogram generation and spectrogram-to-speech generation, respectively; 4

% On all the benchmark datasets, our TranSpeech achieves the best translation quality. 

% the acoustic variation in original self-supervised representations could be hard to model, while speech analysis and bilateral perturbation 

\begin{table}
    \centering
    \small
    \vspace{-10mm}
    % \hspace{-4mm}
    \caption{\textbf{Translation quality (BLEU scores ($\uparrow$)) and inference speed (frame/second ($\uparrow$)) comparison with baseline systems.} We set beam size to $5$ in autoregressive decoding, and apply 5 iterative cycles in NAR naive decoding. \dag: In this work, we remove the auxiliary task (e.g., source and target CTC, auto-encoding) in training the S2ST system for simplification. Though the S2ST system can be further improved with the auxiliary task, this is beyond our focus. BiP: Bilateral Perturbation; NPD: noisy parallel decoding; b: length beam in NAR decoding. }
    \vspace{4mm}
    \scalebox{1.00}{
    \begin{tabular}{c|l|c|c|c|c|c|c}
    \toprule
    ID & Model                                      & BiP                 & Fr-En     & En-Fr      & En-Es     & Speed    & Speedup \\
    \midrule        
    \multicolumn{6}{l}{\bfseries Autoregressive models}    \\
    \midrule 
    1  & Basic Transformer~\citep{lee2021direct}\dag           &  \XSolidBrush  & 15.44    & 15.28    &  10.07    & \multirow{2}{*}{870}    & \multirow{2}{*}{1.00$\times$} \\
    2  & Basic Norm Transformer~\citep{lee2021textless}\dag       & \XSolidBrush   & 15.81    &  15.93   &  12.98 &   &   \\ \cline{7-8}
    3   & Basic Conformer                     &  \XSolidBrush       & 18.02    &  17.07   &  13.75    & \multirow{2}{*}{895}   & \multirow{2}{*}{1.02$\times$}   \\ 
    4   & Basic Conformer          & \CheckmarkBold      &  \textbf{22.39}   &  \textbf{19.65}    & \textbf{14.94}     &  & \\
    \midrule  
    \multicolumn{6}{l}{\bfseries Non-autoregressive models with naive decoding}    \\  
    \midrule            
    5   & TranSpeech - Distill                 & \XSolidBrush           &  14.86   &  14.12  & 10.27   & \multirow{3}{*}{\bfseries 9610}    & \multirow{3}{*}{\bfseries 11.04$\times$} \\
    6   & Transpeech - Distill                 & \CheckmarkBold         & 16.23    & 15.9    & 10.94   &    &  \\
    7   & TranSpeech                           & \CheckmarkBold         & \textbf{17.24}   &  \textbf{16.3}   & \textbf{11.79}   &   & \\
    \midrule  
    \multicolumn{6}{l}{\bfseries Non-autoregressive models with advanced decoding}    \\
    \midrule 
    8   & TranSpeech (iter=15)                & \CheckmarkBold  &   18.03    & 16.97     & 12.62    & 4651  & 5.34$\times$ \\
    9   & TranSpeech (iter=15 + b=15)         & \CheckmarkBold  &   18.10    & 17.05     & 12.70     & 2394   & 2.75$\times$  \\
    10   & TranSpeech (iter=15 + b=15 + NPD)  & \CheckmarkBold  &  \textbf{18.39}     & \textbf{17.50}     & \textbf{12.77}     & 2208  & 2.53$\times$ \\
    \midrule    
    \multicolumn{6}{l}{\bfseries Cascaded systems}  \\
    \midrule
    11  & S2T + TTS                           & \textbf{/}      &  27.17      &  34.85    &  32.86     & \textbf{/} & \textbf{/} \\
    12  & Direct ASR                          & \textbf{/}      &  71.61      &  50.92    &  68.75      &\textbf{/}  &\textbf{/} \\
    13  & Direct TTS                          & \textbf{/}      &  82.41      &  76.87     & 83.69      &\textbf{/}  &\textbf{/} \\
    \bottomrule
    \end{tabular}}
    \label{table1}
    \end{table}

\subsection{Translation Accuracy and Speech Naturalness}
\begin{figure*}[ht]
    \centering
    \small
    \vspace{-4mm}
    \subfigure[Translation latency]
    {
        \label{fig:latency}
      \includegraphics[width=0.45\textwidth]{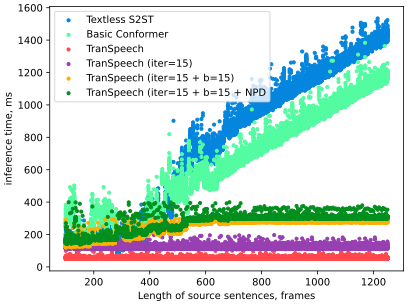}
    }
    \vspace{-2mm}
    \subfigure[Performance-speed trade-off.]
    {
      \label{fig:trade_off}
      \includegraphics[width=0.45\textwidth]{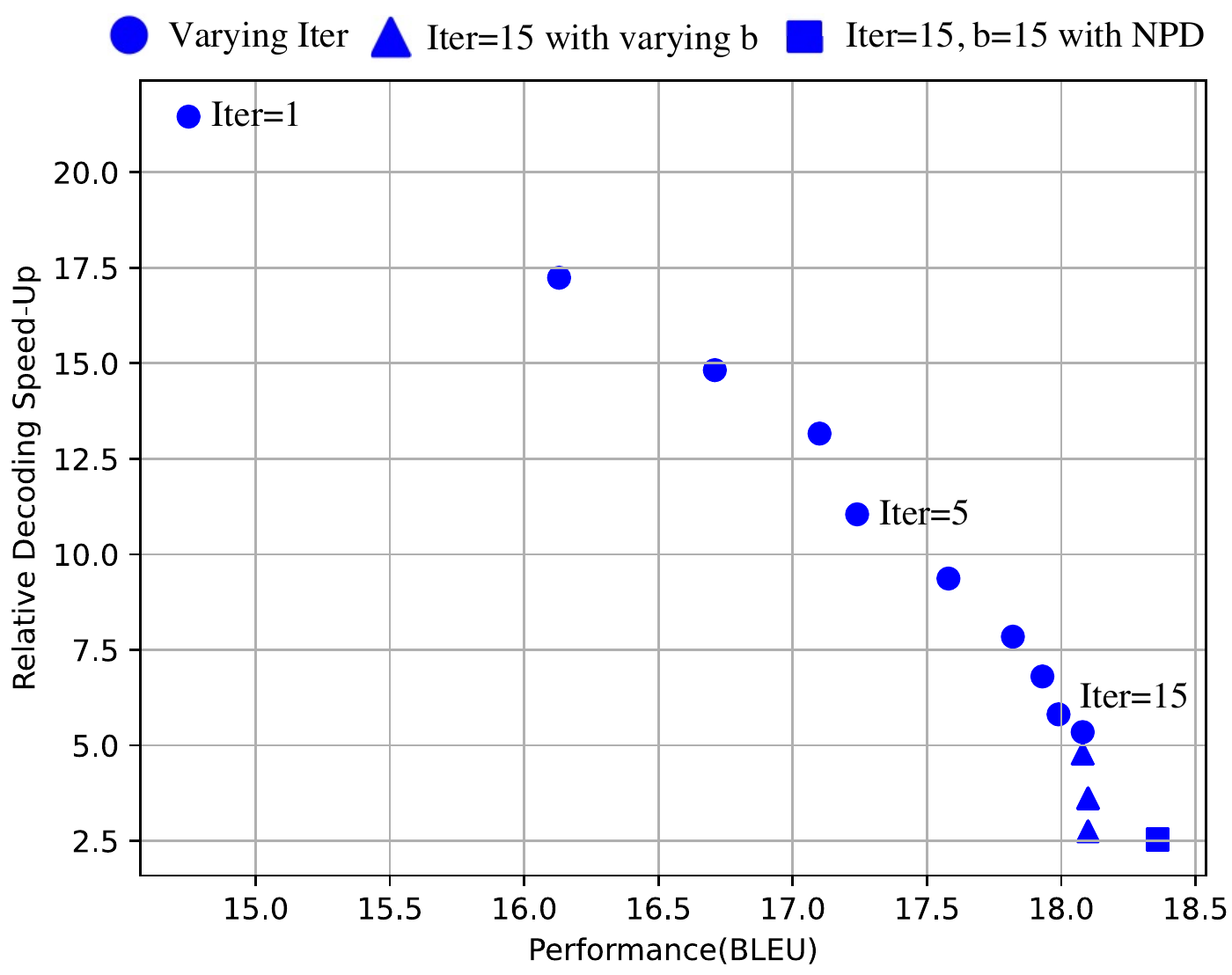}
    }
    \vspace{-1mm}
    \caption{The translation latency is computed as the time to decode the n-frame speech sample, averaged over the test set using 1 NVIDIA V100. b: length beam. NPD: noisy parallel decoding.} 
    \vspace{-3mm}
    \end{figure*}

Table~\ref{table1} summarizes the translation accuracy and inference latency among all systems, and we have the following observations: 1) \textbf{Bilateral perturbation (3 vs. 4)} improves S2ST performance by a large margin of 2.9 BLEU points. The proposed techniques address acoustic multimodality by disentangling the acoustic information and learning linguistic representation given speech samples, which produce more deterministic targets in speech-to-unit translation. 2) \textbf{Conformer architecture (2 vs. 3)} shows a 2.2 BLEU gain of translation accuracy. It combines convolution neural networks and transformers as joint architecture, exhibiting outperformed ability in learning local and global dependencies of an audio. 3) \textbf{Knowledge distillation (6 vs. 7)} is demonstrated to alleviate the linguistic multimodality where training on the distillation corpus provides a distinct promotion of around 1 BLEU points. For speech quality, we attach evaluation in Appendix~\ref{speech_quality}. 
% and compare systems in various benchmarks
When considering the \textbf{speed-performance trade-off in the NAR unit decoder}, we find that more iterative cycles (7 vs. 8), or advanced decoding methods (e.g., length beam (8 vs. 9) and noisy parallel decoding (9 vs. 10)) further lead to an improvement of translation accuracy, trading up to 1.5 BLEU points during decoding. In comparison with baseline systems, TranSpeech yields the highest BLEU scores than the best publicly-available direct S2ST baselines (2 vs. 6) by a considerable margin; in fact, only 2 mask-predict iterations (see Figure~\ref{fig:trade_off}) are necessary for achieving a new SOTA on textless S2ST.

% 2) Across the three language pairs we use, the NAT performs between 3-4 BLEU points worse than its autoregressive teacher (3 vs. 5), with part or all of this gap addressed by the use of advanced non-autoregressive decoding, including multiple iteration, length beam and noisy parallel decoding.

% Besides, compared with S2T+TTS cascaded systems (11) that uses extra ASR models for converting speech to text for training the translation model, a gap could still be witnessed under the textless speech-to-speech translation setting without auxiliary task. Though the S2T system can be further improved with text-based pre-training, this is beyond our scope. the approach has the potential to be adopted for building direct S2ST systems for written languages as well to enable faster inference for S2ST.

\subsection{Decoding Speed}
%   The autoregressive model has latency linear in the decoding length, while the latency of the TranSpeech is nearly constant for typical lengths. When using advanced decoding (more refinement iterations, length beam, and noisy parallel decoding), the level of parallelism is enough to more than saturate the GPU, leading again to linear latencies.

We visualize the relationship between the translation latency and the length of input speech in Figure~\ref{fig:latency}. As can be seen, the autoregressive baselines have a latency linear in the decoding length. At the same time, NAR TranSpeech is nearly constant for typical lengths, even with multiple cycles of mask-predict iterative refinement. We further illustrate the versatile speed-performance trade-off for NAR decoding in Figure~\ref{fig:trade_off}. TranSpeech enables a speedup up to 21.4x compared to the autoregressive baseline. On the other, it could alternatively retain the highest quality with BELU 18.39 while gaining a 253\% speedup.

% ; 2) TranSpeech with non-autoregressive decoding achieves 11x speedup compared to autoregressive models and enjoy the parallelism. 

% we can translate over 15 times faster than autoregressive baseline and yields the outperformed BLEU scores,
    
% \begin{figure}[h]
%     \centering
%     \vspace{-8mm}
%     \includegraphics[width=0.45\textwidth]{Figures/latency.png}
%     \vspace{-1mm}
%     \caption{The translation latency, which is computed as the time to decode a n-frame speech sample, averaged over the whole test set using 1 NVIDIA Tesla V100.} 
%     \vspace{-2mm}
%     \label{fig:latency}
%   \end{figure}
\vspace{-3mm}
\subsection{Case Study}

\begin{table}[]
    \centering
    \small
    \vspace{-4mm}
    \caption{\textbf{Two examples comparing translations produced by TranSpeech and baseline models.} We use the bond fonts to indicate the the issue of \textbf{noisy and incomplete translation.}}
    \vspace{2mm}
    \begin{tabular}{l|l}
    \toprule
    Source:      & l'origine de la rue est liée à la construction de la place rihour.                \\
    Target:      & the origin of the street is linked to the construction of rihour square.           \\
    Basic Conformer:          & the origin of the street is linked to the construction of \bf{the}.                \\
    TranSpeech:         & th origin of the \textbf{seti} is linked to the construction of the \textbf{rear}. \\
    TranSpeech+BiP:     & the origin of the street is linked to the construction of the \bf{ark}.            \\
    TranSpeech+BiP+Advanced: & the origin of the street is linked to the construction of the work.                \\
    \midrule
    Source:      & il participe aux activités du patronage laïque et des pionniers de saint-ouen. \\
    Target:      & he participates in the secular patronage and pioneer activities of saint ouen. \\
    Basic Conformer:          & he participated in the activities of the late patronage a \textbf{d} see.  \\
    TranSpeech:         & he \bf{takes in the patronage activities in of saint}. \\
    TranSpeech+BiP:     & he participated in the activities of the lake patronage and \textbf{say pointing} \\
    TranSpeech+BiP+Advanced: & he participated in the activities of the wake patronage and saint pioneers \\
    \bottomrule
    \end{tabular}
    \label{table4}
\vspace{-4mm}
    \end{table}

    % including the source sentence, the target reference (i.e., the ground-truth translation), the translation given by autoregressive baselines and TranSpeech
We present several translation examples sampled from the Fr-En language pair in Table~\ref{table4}, and have the following findings: 1) Models trained with original units suffer severely from the issue of \textit{noisy and incomplete translation} due to the indeterministic training targets, while with the bilateral perturbation brought in, this multimodal issue is largely alleviated; 2) the advanced decoding methods lead to a distinct improvement in translation accuracy. As can be seen, the results produced by the TranSpeech with advanced decoding (more iterations and NPD), while of a similar quality to those produced by the autoregressive basic conformer, are noticeably more literal. 

\subsection{Ablation Study}

\begin{wraptable}{r}{9cm}
    \centering
    \small
    \vspace{-8mm}
    \caption{\textbf{Ablation study results.} SN: style normalization; IE: information enhancement; PE: positional encoding.}
    \vspace{2mm}
    \begin{tabular}{c|l|c|ccc}
    \toprule
    ID & Model            & PE                      & Fr-En & En-Fr & En-Es \\
    \midrule
    1  & Basic Conformer  & Relative                & 18.02   & 17.07   &  13.75  \\
    2  & Basic Conformer + IE  & Relative           & 21.98  &  19.60   & 14.91       \\
    3  & Basic Conformer + SN  & Relative           & 21.54  & 18.53   & 13.97        \\
    4  & Basic Conformer  & Absolute                & 17.23  & 16.19   & 13.06      \\
    \bottomrule
    \end{tabular}
    % \vspace{-3mm}
\label{table3}
\end{wraptable}

We conduct ablation studies to demonstrate the effectiveness of several detailed designs in this work, including the bilateral perturbation and the conformer architecture in TranSpeech. The results have been presented in Table~\ref{table3}, and we have the following observations: 1) Style normalization and information enhancement in bilateral perturbation both demonstrate a performance gain, and they work in a joint effort to learn deterministic representations, leading to improvements in translation accuracy. 2) Replacing the relative positional encoding in the self-attention layer by the vanilla one~\citep{vaswani2017attention} witnesses a distinct degradation in translation accuracy, demonstrating the outperformed capability of modeling both local and global audio dependencies brought by architecture designs.

\section{Conclusion} \label{con}

In this work, we propose TranSpeech, a speech-to-speech translation model with bilateral perturbation. To tackle the acoustic multimodal issue in S2ST, the bilateral perturbation, which included style normalization and information enhancement, had been proposed to learn only the linguistic information from acoustic-variant speech samples. It assisted in generating deterministic representation agnostic to acoustic conditions, significantly reducing the acoustic multimodality and making it possible for non-autoregressive (NAR) generation. As such, we further stepped forward and became the first to establish a NAR S2ST technique. TranSpeech took full advantage of parallelism and leveraged the mask-predict algorithm to generate results in a constant number of iterations. To address linguistic multimodality, we applied knowledge distillation by constructing a less noisy sampled translation corpus. Experimental results demonstrated that BiP yields an improvement of 2.9 BLEU on average compared with a baseline textless S2ST model. Moreover, TranSpeech showed a significant improvement in inference latency, which required as few as 2 iterations to generate outperformed samples, enabling a sampling speed of up to 21.4x faster than the autoregressive baseline. We envisage that our work will serve as a basis for future textless S2ST studies.

\section*{Acknowledgements}
This work was supported in part by the National Natural Science Foundation of China under Grant No. 62222211, National Key R\&D Program of China under Grant No.2020YFC0832505, Zhejiang Electric Power Co., Ltd. Science and Technology Project No.5211YF22006 and Yiwise.

\bibliography{iclr2023_conference}

\begin{thebibliography}{64}
\providecommand{\natexlab}[1]{#1}
\providecommand{\url}[1]{\texttt{#1}}
\expandafter\ifx\csname urlstyle\endcsname\relax
  \providecommand{\doi}[1]{doi: #1}\else
  \providecommand{\doi}{doi: \begingroup \urlstyle{rm}\Url}\fi

\bibitem[Baevski et~al.(2019)Baevski, Auli, and
  Mohamed]{baevski2019effectiveness}
Alexei Baevski, Michael Auli, and Abdelrahman Mohamed.
\newblock Effectiveness of self-supervised pre-training for speech recognition.
\newblock \emph{arXiv preprint arXiv:1911.03912}, 2019.

\bibitem[Baevski et~al.(2020)Baevski, Zhou, Mohamed, and
  Auli]{baevski2020wav2vec}
Alexei Baevski, Yuhao Zhou, Abdelrahman Mohamed, and Michael Auli.
\newblock wav2vec 2.0: A framework for self-supervised learning of speech
  representations.
\newblock \emph{Advances in Neural Information Processing Systems},
  33:\penalty0 12449--12460, 2020.

\bibitem[Chen et~al.(2020)Chen, Watanabe, Villalba, {\.Z}elasko, and
  Dehak]{chen2020non}
Nanxin Chen, Shinji Watanabe, Jes{\'u}s Villalba, Piotr {\.Z}elasko, and Najim
  Dehak.
\newblock Non-autoregressive transformer for speech recognition.
\newblock \emph{IEEE Signal Processing Letters}, 28:\penalty0 121--125, 2020.

\bibitem[Chen et~al.(2021)Chen, Wu, Chen, Wu, Li, Yoshioka, Wang, Liu, and
  Zhou]{chen2021continuous}
Sanyuan Chen, Yu~Wu, Zhuo Chen, Jian Wu, Jinyu Li, Takuya Yoshioka, Chengyi
  Wang, Shujie Liu, and Ming Zhou.
\newblock Continuous speech separation with conformer.
\newblock In \emph{ICASSP 2021-2021 IEEE International Conference on Acoustics,
  Speech and Signal Processing (ICASSP)}, pp.\  5749--5753. IEEE, 2021.

\bibitem[Choi et~al.(2021)Choi, Lee, Kim, Lee, Heo, and Lee]{choi2021neural}
Hyeong-Seok Choi, Juheon Lee, Wansoo Kim, Jie Lee, Hoon Heo, and Kyogu Lee.
\newblock Neural analysis and synthesis: Reconstructing speech from
  self-supervised representations.
\newblock \emph{Advances in Neural Information Processing Systems}, 34, 2021.

\bibitem[Chorowski et~al.(2019)Chorowski, Weiss, Bengio, and Van
  Den~Oord]{chorowski2019unsupervised}
Jan Chorowski, Ron~J Weiss, Samy Bengio, and A{\"a}ron Van Den~Oord.
\newblock Unsupervised speech representation learning using wavenet
  autoencoders.
\newblock \emph{IEEE/ACM transactions on audio, speech, and language
  processing}, 27\penalty0 (12):\penalty0 2041--2053, 2019.

\bibitem[Chung et~al.(2019)Chung, Hsu, Tang, and Glass]{chung2019unsupervised}
Yu-An Chung, Wei-Ning Hsu, Hao Tang, and James Glass.
\newblock An unsupervised autoregressive model for speech representation
  learning.
\newblock \emph{arXiv preprint arXiv:1904.03240}, 2019.

\bibitem[Cui et~al.(2021)Cui, Ren, Liu, Chen, Huang, Lei, and
  Zhao]{cui2021emovie}
Chenye Cui, Yi~Ren, Jinglin Liu, Feiyang Chen, Rongjie Huang, Ming Lei, and
  Zhou Zhao.
\newblock Emovie: A mandarin emotion speech dataset with a simple emotional
  text-to-speech model.
\newblock \emph{arXiv preprint arXiv:2106.09317}, 2021.

\bibitem[Cui et~al.(2022)Cui, Ren, Liu, Huang, and Zhao]{cui2022varietysound}
Chenye Cui, Yi~Ren, Jinglin Liu, Rongjie Huang, and Zhou Zhao.
\newblock Varietysound: Timbre-controllable video to sound generation via
  unsupervised information disentanglement.
\newblock \emph{arXiv preprint arXiv:2211.10666}, 2022.

\bibitem[Dai et~al.(2019)Dai, Yang, Yang, Carbonell, Le, and
  Salakhutdinov]{dai2019transformer}
Zihang Dai, Zhilin Yang, Yiming Yang, Jaime Carbonell, Quoc~V Le, and Ruslan
  Salakhutdinov.
\newblock Transformer-xl: Attentive language models beyond a fixed-length
  context.
\newblock \emph{arXiv preprint arXiv:1901.02860}, 2019.

\bibitem[Devlin et~al.(2018)Devlin, Chang, Lee, and Toutanova]{devlin2018bert}
Jacob Devlin, Ming-Wei Chang, Kenton Lee, and Kristina Toutanova.
\newblock Bert: Pre-training of deep bidirectional transformers for language
  understanding.
\newblock \emph{arXiv preprint arXiv:1810.04805}, 2018.

\bibitem[Gao et~al.(2022)Gao, Ni, Qian, Zhang, Chang, and
  Hasegawa-Johnson]{gao2022wavprompt}
Heting Gao, Junrui Ni, Kaizhi Qian, Yang Zhang, Shiyu Chang, and Mark
  Hasegawa-Johnson.
\newblock Wavprompt: Towards few-shot spoken language understanding with frozen
  language models.
\newblock \emph{arXiv preprint arXiv:2203.15863}, 2022.

\bibitem[Ghazvininejad et~al.(2019)Ghazvininejad, Levy, Liu, and
  Zettlemoyer]{ghazvininejad2019mask}
Marjan Ghazvininejad, Omer Levy, Yinhan Liu, and Luke Zettlemoyer.
\newblock Mask-predict: Parallel decoding of conditional masked language
  models.
\newblock \emph{arXiv preprint arXiv:1904.09324}, 2019.

\bibitem[Gu et~al.(2017)Gu, Bradbury, Xiong, Li, and Socher]{gu2017non}
Jiatao Gu, James Bradbury, Caiming Xiong, Victor~OK Li, and Richard Socher.
\newblock Non-autoregressive neural machine translation.
\newblock \emph{arXiv preprint arXiv:1711.02281}, 2017.

\bibitem[Gu et~al.(2019)Gu, Wang, and Zhao]{gu2019levenshtein}
Jiatao Gu, Changhan Wang, and Junbo Zhao.
\newblock Levenshtein transformer.
\newblock \emph{Advances in Neural Information Processing Systems}, 32, 2019.

\bibitem[Gulati et~al.(2020)Gulati, Qin, Chiu, Parmar, Zhang, Yu, Han, Wang,
  Zhang, Wu, et~al.]{gulati2020conformer}
Anmol Gulati, James Qin, Chung-Cheng Chiu, Niki Parmar, Yu~Zhang, Jiahui Yu,
  Wei Han, Shibo Wang, Zhengdong Zhang, Yonghui Wu, et~al.
\newblock Conformer: Convolution-augmented transformer for speech recognition.
\newblock \emph{arXiv preprint arXiv:2005.08100}, 2020.

\bibitem[Guo et~al.(2021)Guo, Boyer, Chang, Hayashi, Higuchi, Inaguma, Kamo,
  Li, Garcia-Romero, Shi, et~al.]{guo2021recent}
Pengcheng Guo, Florian Boyer, Xuankai Chang, Tomoki Hayashi, Yosuke Higuchi,
  Hirofumi Inaguma, Naoyuki Kamo, Chenda Li, Daniel Garcia-Romero, Jiatong Shi,
  et~al.
\newblock Recent developments on espnet toolkit boosted by conformer.
\newblock In \emph{ICASSP 2021-2021 IEEE International Conference on Acoustics,
  Speech and Signal Processing (ICASSP)}, pp.\  5874--5878. IEEE, 2021.

\bibitem[Hsu et~al.(2020)Hsu, Harwath, Song, and Glass]{hsu2020text}
Wei-Ning Hsu, David Harwath, Christopher Song, and James Glass.
\newblock Text-free image-to-speech synthesis using learned segmental units.
\newblock \emph{arXiv preprint arXiv:2012.15454}, 2020.

\bibitem[Hsu et~al.(2021)Hsu, Bolte, Tsai, Lakhotia, Salakhutdinov, and
  Mohamed]{hsu2021hubert}
Wei-Ning Hsu, Benjamin Bolte, Yao-Hung~Hubert Tsai, Kushal Lakhotia, Ruslan
  Salakhutdinov, and Abdelrahman Mohamed.
\newblock Hubert: Self-supervised speech representation learning by masked
  prediction of hidden units.
\newblock \emph{IEEE/ACM Transactions on Audio, Speech, and Language
  Processing}, 29:\penalty0 3451--3460, 2021.

\bibitem[Huang et~al.(2021)Huang, Chen, Ren, Liu, Cui, and
  Zhao]{huang2021multi}
Rongjie Huang, Feiyang Chen, Yi~Ren, Jinglin Liu, Chenye Cui, and Zhou Zhao.
\newblock Multi-singer: Fast multi-singer singing voice vocoder with a
  large-scale corpus.
\newblock In \emph{Proceedings of the 29th ACM International Conference on
  Multimedia}, pp.\  3945--3954, 2021.

\bibitem[Huang et~al.(2022{\natexlab{a}})Huang, Cui, Chen, Ren, Liu, Zhao,
  Huai, and Wang]{huang2022singgan}
Rongjie Huang, Chenye Cui, Feiyang Chen, Yi~Ren, Jinglin Liu, Zhou Zhao,
  Baoxing Huai, and Zhefeng Wang.
\newblock Singgan: Generative adversarial network for high-fidelity singing
  voice generation.
\newblock In \emph{Proceedings of the 30th ACM International Conference on
  Multimedia}, pp.\  2525--2535, 2022{\natexlab{a}}.

\bibitem[Huang et~al.(2022{\natexlab{b}})Huang, Lam, Wang, Su, Yu, Ren, and
  Zhao]{huang2022fastdiff}
Rongjie Huang, Max~WY Lam, Jun Wang, Dan Su, Dong Yu, Yi~Ren, and Zhou Zhao.
\newblock Fastdiff: A fast conditional diffusion model for high-quality speech
  synthesis.
\newblock \emph{arXiv preprint arXiv:2204.09934}, 2022{\natexlab{b}}.

\bibitem[Huang et~al.(2022{\natexlab{c}})Huang, Ren, Liu, Cui, and
  Zhao]{huang2022generspeech}
Rongjie Huang, Yi~Ren, Jinglin Liu, Chenye Cui, and Zhou Zhao.
\newblock Generspeech: Towards style transfer for generalizable out-of-domain
  text-to-speech synthesis.
\newblock \emph{arXiv preprint arXiv:2205.07211}, 2022{\natexlab{c}}.

\bibitem[Huang et~al.(2022{\natexlab{d}})Huang, Zhao, Liu, Liu, Cui, and
  Ren]{huang2022prodiff}
Rongjie Huang, Zhou Zhao, Huadai Liu, Jinglin Liu, Chenye Cui, and Yi~Ren.
\newblock Prodiff: Progressive fast diffusion model for high-quality
  text-to-speech.
\newblock \emph{arXiv preprint arXiv:2207.06389}, 2022{\natexlab{d}}.

\bibitem[Jia et~al.(2019)Jia, Weiss, Biadsy, Macherey, Johnson, Chen, and
  Wu]{jia2019direct}
Ye~Jia, Ron~J Weiss, Fadi Biadsy, Wolfgang Macherey, Melvin Johnson, Zhifeng
  Chen, and Yonghui Wu.
\newblock Direct speech-to-speech translation with a sequence-to-sequence
  model.
\newblock \emph{arXiv preprint arXiv:1904.06037}, 2019.

\bibitem[Jia et~al.(2021)Jia, Ramanovich, Remez, and
  Pomerantz]{jia2021translatotron}
Ye~Jia, Michelle~Tadmor Ramanovich, Tal Remez, and Roi Pomerantz.
\newblock Translatotron 2: Robust direct speech-to-speech translation.
\newblock \emph{arXiv preprint arXiv:2107.08661}, 2021.

\bibitem[Jia et~al.(2022)Jia, Ramanovich, Wang, and Zen]{jia2022cvss}
Ye~Jia, Michelle~Tadmor Ramanovich, Quan Wang, and Heiga Zen.
\newblock Cvss corpus and massively multilingual speech-to-speech translation.
\newblock \emph{arXiv preprint arXiv:2201.03713}, 2022.

\bibitem[Kong et~al.(2020)Kong, Kim, and Bae]{kong2020hifi}
Jungil Kong, Jaehyeon Kim, and Jaekyoung Bae.
\newblock Hifi-gan: Generative adversarial networks for efficient and high
  fidelity speech synthesis.
\newblock \emph{Advances in Neural Information Processing Systems},
  33:\penalty0 17022--17033, 2020.

\bibitem[Lam et~al.(2021)Lam, Wang, Huang, Su, and Yu]{lam2021bilateral}
Max~WY Lam, Jun Wang, Rongjie Huang, Dan Su, and Dong Yu.
\newblock Bilateral denoising diffusion models.
\newblock \emph{arXiv preprint arXiv:2108.11514}, 2021.

\bibitem[Lavie et~al.(1997)Lavie, Waibel, Levin, Finke, Gates, Gavalda,
  Zeppenfeld, and Zhan]{lavie1997janus}
Alon Lavie, Alex Waibel, Lori Levin, Michael Finke, Donna Gates, Marsal
  Gavalda, Torsten Zeppenfeld, and Puming Zhan.
\newblock Janus-iii: Speech-to-speech translation in multiple languages.
\newblock In \emph{1997 IEEE International Conference on Acoustics, Speech, and
  Signal Processing}, volume~1, pp.\  99--102. IEEE, 1997.

\bibitem[Lee et~al.(2021{\natexlab{a}})Lee, Chen, Wang, Gu, Ma, Polyak, Adi,
  He, Tang, Pino, et~al.]{lee2021direct}
Ann Lee, Peng-Jen Chen, Changhan Wang, Jiatao Gu, Xutai Ma, Adam Polyak, Yossi
  Adi, Qing He, Yun Tang, Juan Pino, et~al.
\newblock Direct speech-to-speech translation with discrete units.
\newblock \emph{arXiv preprint arXiv:2107.05604}, 2021{\natexlab{a}}.

\bibitem[Lee et~al.(2021{\natexlab{b}})Lee, Gong, Duquenne, Schwenk, Chen,
  Wang, Popuri, Pino, Gu, and Hsu]{lee2021textless}
Ann Lee, Hongyu Gong, Paul-Ambroise Duquenne, Holger Schwenk, Peng-Jen Chen,
  Changhan Wang, Sravya Popuri, Juan Pino, Jiatao Gu, and Wei-Ning Hsu.
\newblock Textless speech-to-speech translation on real data.
\newblock \emph{arXiv preprint arXiv:2112.08352}, 2021{\natexlab{b}}.

\bibitem[Lin et~al.(2021)Lin, Zhao, Li, Liu, Zhang, Zeng, and
  He]{lin2021simullr}
Zhijie Lin, Zhou Zhao, Haoyuan Li, Jinglin Liu, Meng Zhang, Xingshan Zeng, and
  Xiaofei He.
\newblock Simullr: Simultaneous lip reading transducer with attention-guided
  adaptive memory.
\newblock In \emph{Proceedings of the 29th ACM International Conference on
  Multimedia}, pp.\  1359--1367, 2021.

\bibitem[Nakamura et~al.(2006)Nakamura, Markov, Nakaiwa, Kikui, Kawai,
  Jitsuhiro, Zhang, Yamamoto, Sumita, and Yamamoto]{nakamura2006atr}
Satoshi Nakamura, Konstantin Markov, Hiromi Nakaiwa, Gen-ichiro Kikui, Hisashi
  Kawai, Takatoshi Jitsuhiro, J-S Zhang, Hirofumi Yamamoto, Eiichiro Sumita,
  and Seiichi Yamamoto.
\newblock The atr multilingual speech-to-speech translation system.
\newblock \emph{IEEE Transactions on Audio, Speech, and Language Processing},
  14\penalty0 (2):\penalty0 365--376, 2006.

\bibitem[Ott et~al.(2019)Ott, Edunov, Baevski, Fan, Gross, Ng, Grangier, and
  Auli]{ott2019fairseq}
Myle Ott, Sergey Edunov, Alexei Baevski, Angela Fan, Sam Gross, Nathan Ng,
  David Grangier, and Michael Auli.
\newblock fairseq: A fast, extensible toolkit for sequence modeling.
\newblock \emph{arXiv preprint arXiv:1904.01038}, 2019.

\bibitem[Papineni et~al.(2002)Papineni, Roukos, Ward, and
  Zhu]{papineni2002bleu}
Kishore Papineni, Salim Roukos, Todd Ward, and Wei-Jing Zhu.
\newblock Bleu: a method for automatic evaluation of machine translation.
\newblock In \emph{Proceedings of the 40th annual meeting of the Association
  for Computational Linguistics}, pp.\  311--318, 2002.

\bibitem[Polyak \& Wolf(2019)Polyak and Wolf]{polyak2019attention}
Adam Polyak and Lior Wolf.
\newblock Attention-based wavenet autoencoder for universal voice conversion.
\newblock In \emph{ICASSP 2019-2019 IEEE International Conference on Acoustics,
  Speech and Signal Processing (ICASSP)}, pp.\  6800--6804. IEEE, 2019.

\bibitem[Polyak et~al.(2021)Polyak, Adi, Copet, Kharitonov, Lakhotia, Hsu,
  Mohamed, and Dupoux]{polyak2021speech}
Adam Polyak, Yossi Adi, Jade Copet, Eugene Kharitonov, Kushal Lakhotia,
  Wei-Ning Hsu, Abdelrahman Mohamed, and Emmanuel Dupoux.
\newblock Speech resynthesis from discrete disentangled self-supervised
  representations.
\newblock \emph{arXiv preprint arXiv:2104.00355}, 2021.

\bibitem[Qian et~al.(2020)Qian, Zhang, Chang, Hasegawa-Johnson, and
  Cox]{qian2020unsupervised}
Kaizhi Qian, Yang Zhang, Shiyu Chang, Mark Hasegawa-Johnson, and David Cox.
\newblock Unsupervised speech decomposition via triple information bottleneck.
\newblock In \emph{International Conference on Machine Learning}, pp.\
  7836--7846. PMLR, 2020.

\bibitem[Qian et~al.(2021)Qian, Zhang, Chang, Xiong, Gan, Cox, and
  Hasegawa-Johnson]{qian2021global}
Kaizhi Qian, Yang Zhang, Shiyu Chang, Jinjun Xiong, Chuang Gan, David Cox, and
  Mark Hasegawa-Johnson.
\newblock Global prosody style transfer without text transcriptions.
\newblock In \emph{International Conference on Machine Learning}, pp.\
  8650--8660. PMLR, 2021.

\bibitem[Ren et~al.(2020)Ren, Hu, Tan, Qin, Zhao, Zhao, and
  Liu]{ren2020fastspeech}
Yi~Ren, Chenxu Hu, Xu~Tan, Tao Qin, Sheng Zhao, Zhou Zhao, and Tie-Yan Liu.
\newblock Fastspeech 2: Fast and high-quality end-to-end text to speech.
\newblock \emph{arXiv preprint arXiv:2006.04558}, 2020.

\bibitem[Vaswani et~al.(2017)Vaswani, Shazeer, Parmar, Uszkoreit, Jones, Gomez,
  Kaiser, and Polosukhin]{vaswani2017attention}
Ashish Vaswani, Noam Shazeer, Niki Parmar, Jakob Uszkoreit, Llion Jones,
  Aidan~N Gomez, {\L}ukasz Kaiser, and Illia Polosukhin.
\newblock Attention is all you need.
\newblock \emph{Advances in neural information processing systems}, 30, 2017.

\bibitem[Wahlster(2013)]{wahlster2013verbmobil}
Wolfgang Wahlster.
\newblock \emph{Verbmobil: foundations of speech-to-speech translation}.
\newblock Springer Science \& Business Media, 2013.

\bibitem[Wang et~al.(2020{\natexlab{a}})Wang, Tang, Ma, Wu, Okhonko, and
  Pino]{wang2020fairseqs2t}
Changhan Wang, Yun Tang, Xutai Ma, Anne Wu, Dmytro Okhonko, and Juan Pino.
\newblock fairseq s2t: Fast speech-to-text modeling with fairseq.
\newblock In \emph{Proceedings of the 2020 Conference of the Asian Chapter of
  the Association for Computational Linguistics (AACL): System Demonstrations},
  2020{\natexlab{a}}.

\bibitem[Wang et~al.(2020{\natexlab{b}})Wang, Wu, and Pino]{wang2020covost}
Changhan Wang, Anne Wu, and Juan Pino.
\newblock Covost 2 and massively multilingual speech-to-text translation.
\newblock \emph{arXiv preprint arXiv:2007.10310}, 2020{\natexlab{b}}.

\bibitem[Wang et~al.(2019)Wang, Tian, He, Qin, Zhai, and Liu]{wang2019non}
Yiren Wang, Fei Tian, Di~He, Tao Qin, ChengXiang Zhai, and Tie-Yan Liu.
\newblock Non-autoregressive machine translation with auxiliary regularization.
\newblock In \emph{Proceedings of the AAAI conference on artificial
  intelligence}, volume~33, pp.\  5377--5384, 2019.

\bibitem[Xia et~al.(2022)Xia, Zhao, Ye, Zhao, Li, and Ren]{xia2022video}
Yan Xia, Zhou Zhao, Shangwei Ye, Yang Zhao, Haoyuan Li, and Yi~Ren.
\newblock Video-guided curriculum learning for spoken video grounding.
\newblock In \emph{Proceedings of the 30th ACM International Conference on
  Multimedia}, pp.\  5191--5200, 2022.

\bibitem[Yang et~al.(2019)Yang, Liu, and Zou]{yang2019non}
Bang Yang, Fenglin Liu, and Yuexian Zou.
\newblock Non-autoregressive video captioning with iterative refinement.
\newblock 2019.

\bibitem[Yang et~al.(2021)Yang, Tsai, and Chen]{yang2021voice2series}
Chao-Han~Huck Yang, Yun-Yun Tsai, and Pin-Yu Chen.
\newblock Voice2series: Reprogramming acoustic models for time series
  classification.
\newblock In \emph{International Conference on Machine Learning}, pp.\
  11808--11819. PMLR, 2021.

\bibitem[Yang et~al.(2022{\natexlab{a}})Yang, Liu, Yu, Wang, Weng, and
  Zou]{yang2022norespeech}
Dongchao Yang, Songxiang Liu, Jianwei Yu, Helin Wang, Chao Weng, and Yuexian
  Zou.
\newblock Norespeech: Knowledge distillation based conditional diffusion model
  for noise-robust expressive tts.
\newblock \emph{arXiv preprint arXiv:2211.02448}, 2022{\natexlab{a}}.

\bibitem[Yang et~al.(2022{\natexlab{b}})Yang, Yu, Wang, Wang, Weng, Zou, and
  Yu]{yang2022diffsound}
Dongchao Yang, Jianwei Yu, Helin Wang, Wen Wang, Chao Weng, Yuexian Zou, and
  Dong Yu.
\newblock Diffsound: Discrete diffusion model for text-to-sound generation.
\newblock \emph{arXiv preprint arXiv:2207.09983}, 2022{\natexlab{b}}.

\bibitem[Yang et~al.(2023)Yang, Liu, Huang, Lei, Weng, Meng, and
  Yu]{yang2023instructtts}
Dongchao Yang, Songxiang Liu, Rongjie Huang, Guangzhi Lei, Chao Weng, Helen
  Meng, and Dong Yu.
\newblock Instructtts: Modelling expressive tts in discrete latent space with
  natural language style prompt.
\newblock \emph{arXiv preprint arXiv:2301.13662}, 2023.

\bibitem[Ye et~al.(2022)Ye, Zhao, Ren, and Wu]{ye2022syntaspeech}
Zhenhui Ye, Zhou Zhao, Yi~Ren, and Fei Wu.
\newblock Syntaspeech: Syntax-aware generative adversarial text-to-speech.
\newblock \emph{arXiv preprint arXiv:2204.11792}, 2022.

\bibitem[Ye et~al.(2023)Ye, Jiang, Ren, Liu, He, and Zhao]{ye2023geneface}
Zhenhui Ye, Ziyue Jiang, Yi~Ren, Jinglin Liu, Jinzheng He, and Zhou Zhao.
\newblock Geneface: Generalized and high-fidelity audio-driven 3d talking face
  synthesis.
\newblock \emph{arXiv preprint arXiv:2301.13430}, 2023.

\bibitem[Yen et~al.(2021)Yen, Ku, Yang, Hu, Siniscalchi, Chen, and
  Tsao]{yen2021study}
Hao Yen, Pin-Jui Ku, Chao-Han~Huck Yang, Hu~Hu, Sabato~Marco Siniscalchi,
  Pin-Yu Chen, and Yu~Tsao.
\newblock A study of low-resource speech commands recognition based on
  adversarial reprogramming.
\newblock \emph{arXiv preprint arXiv:2110.03894}, 2021.

\bibitem[Yin et~al.(2021)Yin, Zhao, Liu, Jin, Zhang, Zeng, and
  He]{yin2021simulslt}
Aoxiong Yin, Zhou Zhao, Jinglin Liu, Weike Jin, Meng Zhang, Xingshan Zeng, and
  Xiaofei He.
\newblock Simulslt: End-to-end simultaneous sign language translation.
\newblock In \emph{Proceedings of the 29th ACM International Conference on
  Multimedia}, pp.\  4118--4127, 2021.

\bibitem[Yin et~al.(2022)Yin, Zhao, Jin, Zhang, Zeng, and He]{yin2022mlslt}
Aoxiong Yin, Zhou Zhao, Weike Jin, Meng Zhang, Xingshan Zeng, and Xiaofei He.
\newblock Mlslt: Towards multilingual sign language translation.
\newblock In \emph{Proceedings of the IEEE/CVF Conference on Computer Vision
  and Pattern Recognition}, pp.\  5109--5119, 2022.

\bibitem[Yin et~al.(2023)Yin, Zhong, Tang, Jin, Jin, and Zhao]{GASLT}
Aoxiong Yin, Tianyun Zhong, Li~Tang, Weike Jin, Tao Jin, and Zhou Zhao.
\newblock Gloss attention for gloss-free sign language translation.
\newblock In \emph{{IEEE/CVF} Conference on Computer Vision and Pattern
  Recognition, {CVPR} 2023, Vancouver, Canada, June 17-23, 2023}. {IEEE}, 2023.

\bibitem[Zhang et~al.(2020)Zhang, Tan, Ren, Qin, Zhang, and
  Liu]{zhang2020uwspeech}
Chen Zhang, Xu~Tan, Yi~Ren, Tao Qin, Kejun Zhang, and Tie-Yan Liu.
\newblock Uwspeech: Speech to speech translation for unwritten languages.
\newblock \emph{arXiv preprint arXiv:2006.07926}, 59:\penalty0 132, 2020.

\bibitem[Zhang et~al.()Zhang, Chen, Li, Lyu, Wu, Ding, Shen, and
  Wu]{zhangdense}
Jie Zhang, Chen Chen, Bo~Li, Lingjuan Lyu, Shuang Wu, Shouhong Ding, Chunhua
  Shen, and Chao Wu.
\newblock Dense: Data-free one-shot federated learning.
\newblock In \emph{Advances in Neural Information Processing Systems}.

\bibitem[Zhang et~al.(2022{\natexlab{a}})Zhang, Li, Xu, Wu, Ding, Zhang, and
  Wu]{zhang2022towards}
Jie Zhang, Bo~Li, Jianghe Xu, Shuang Wu, Shouhong Ding, Lei Zhang, and Chao Wu.
\newblock Towards efficient data free black-box adversarial attack.
\newblock In \emph{Proceedings of the IEEE/CVF Conference on Computer Vision
  and Pattern Recognition}, pp.\  15115--15125, 2022{\natexlab{a}}.

\bibitem[Zhang et~al.(2023{\natexlab{a}})Zhang, Li, Chen, Lyu, Wu, Ding, and
  Wu]{zhang2023delving}
Jie Zhang, Bo~Li, Chen Chen, Lingjuan Lyu, Shuang Wu, Shouhong Ding, and Chao
  Wu.
\newblock Delving into the adversarial robustness of federated learning.
\newblock \emph{arXiv preprint arXiv:2302.09479}, 2023{\natexlab{a}}.

\bibitem[Zhang et~al.(2022{\natexlab{b}})Zhang, Zhao, and
  Lin]{zhang2022unsupervised}
Zijian Zhang, Zhou Zhao, and Zhijie Lin.
\newblock Unsupervised representation learning from pre-trained diffusion
  probabilistic models.
\newblock In \emph{Advances in Neural Information Processing Systems},
  2022{\natexlab{b}}.

\bibitem[Zhang et~al.(2023{\natexlab{b}})Zhang, Zhao, Yu, and
  Tian]{zhang2023shiftddpms}
Zijian Zhang, Zhou Zhao, Jun Yu, and Qi~Tian.
\newblock Shiftddpms: Exploring conditional diffusion models by shifting
  diffusion trajectories.
\newblock \emph{arXiv preprint arXiv:2302.02373}, 2023{\natexlab{b}}.

\end{thebibliography}
\bibliographystyle{iclr2023_conference}

\appendix
\clearpage
\begin{center}
    {\bf {\LARGE Appendices}}
\end{center}
\begin{center}
    {\bf {\LARGE TranSpeech: Speech-to-Speech Translation With Bilateral Perturbation} \linebreak}
\end{center}

\section{Related Work} \label{related}

\subsection{Self-Supervised Representation Learning}

There has been an increasing interest in self-supervised learning in the machine learning~\citep{zhang2022towards,lam2021bilateral,zhang2023shiftddpms,zhang2022unsupervised} and multimodal processing community~\citep{xia2022video,zhang2023delving,zhangdense,huang2022fastdiff,huang2022singgan}. Wav2Vec 2.0~\citep{baevski2020wav2vec} trains a convolutional neural network to distinguish true future samples from random distractor samples using a contrastive predictive coding (CPC) loss function. HuBERT~\citep{hsu2021hubert} is trained with a masked prediction with masked continuous audio signals. The majority of self-supervised representation learning models are trained by reconstructing~\citep{chorowski2019unsupervised} or predicting unseen speech signals~\citep{chung2019unsupervised}, which would inevitably include factors unrelated to the linguistic content (i.e., acoustic condition). 
% Although using the self-supervised representations is supposed to get disentangled with most acoustic information, it has been reported~\citep{choi2021neural,hsu2020text} that the acoustic variations still lead to less deterministic results. 

% Contentvec~\citep{qian2022contentvec} designs a random transformation and introduces a teacher-student training mechanism to achieve speaker disentanglement. 

\subsection{Perturbation-based speech reprogramming}
Various approaches that perturb information flow in acoustic models have demonstrated the efficiency in promoting downstream performance: SpeechSplit~\citep{qian2020unsupervised}, AutoPST~\citep{qian2021global}, and NANSY~\citep{choi2021neural} perturb the speech variations during the analysis stage to encourage the synthesis stage to use the supplied more stable representations. Voice2Series~\citep{yang2021voice2series} introduces a novel end-to-end approach that reprograms pre-trained acoustic models for time series classification by input transformation learning and output label mapping. Wavprompt~\citep{gao2022wavprompt} utilizes the pre-trained audio encoder as part of an ASR to convert the speech in the demonstrations into embeddings digestible to the language model. For multi-lingual tuning, ~\citet{yen2021study} propose a novel adversarial reprogramming approach for low-resource spoken command recognition (SCR), which repurposes a pre-trained SCR model to modify the acoustic signals. In this work, we propose the bilateral perturbation technique with style normalization and information enhancement to perturb the acoustic conditions in speech.

\subsection{Non-autoregressive Sequence Generation}

An autoregressive model~\citep{lin2021simullr,yin2021simulslt,yin2022mlslt} takes in a source sequence and then generates target sentences one by one with the causal structure during the inference process. It prevents parallelism during inference, and thus the computational power of GPU cannot be fully exploited. To reduce the inference latency, ~\citep{gu2017non} introduces a non-autoregressive (NAR) transformer-based approach with explicit word fertility, and identifies the multimodality problem of linguistic information between the source and target language. ~\citep{ghazvininejad2019mask} introduced the masked language modeling objective from BERT~\citep{devlin2018bert} to non-autoregressively predict and refine translations. Besides the study of neural machine translation, many works bring NAR model into other sequence-to-sequence tasks~\citep{cui2021emovie,ye2023geneface,huang2022generspeech,yang2022diffsound}, such as video caption~\citep{yang2019non}, speech recognition~\citep{chen2020non} and speech synthesis~\citep{ye2022syntaspeech,huang2022prodiff,yang2023instructtts}. In contrast, we focus on non-autoregressive generation in direct S2ST, which is relatively overlooked.

\section{Model Architectures} \label{appendix:arch}

In this section, we list the model hyper-parameters of TranSpeech in Table \ref{tab:hyperparameters}. 

\begin{table}[h]
\small
\centering
\begin{tabular}{l|c|c}
\toprule
\multicolumn{2}{c|}{Hyperparameter}   & TranSpeech \\ 
\midrule
\multirow{7}{*}{Conformer Encoder} 
&Conv1d Layers                     &    2   \\
&Conv1d Kernel                     &  (5, 5)   \\
&Encoder Block                     &    6     \\
&Encoder Hidden                    &  512      \\                      
&Encoder Attention Heads           &  8      \\    
&Encoder Dropout                   &  0.1   \\ 
\midrule
Length Predictor        &Projection Dim       &  512      \\ 
\midrule
\multirow{6}{*}{Unit Decoder}      
&Unit Dictionary                    &   1000 \\   
&Decoder Block                      &   6 \\    
&Decoder Hidden                      &   512  \\    
&Decoder Attention Headers           &  8   \\       
&Decoder Dropout                     &  0.1  \\     
\bottomrule
\end{tabular}
\vspace{2mm}
\caption{Hyperparameters of TranSpeech.}
\label{tab:hyperparameters}
\end{table}

\section{Impact of indeterministic training target} \label{indeterministic}

% To conclude, the proposed bilateral perturbation technique significantly reduces the multimodal issue, and thus the overall pipeline leverages the less noisy target in training S2UT model and yield better results. 

To visualize the acoustic multimodality and demonstrate the effectiveness of proposed bilateral perturbation, we apply the information bottleneck on acoustic features (i.e., rhythm, pitch, and energy) to create perturbed speech samples $\hat{S_r}, \hat{S_p}, \hat{S_e}$, respectively. We further plot the spectrogram and pitch contours of the original and acoustic-perturbed samples in Figure~\ref{fig:info_perturbation} in Appendix~\ref{vis_per}. The \textbf{unit error rate (UER)} is further adopted as an evaluation matrix to measure the undeterminacy and multimodality according to acoustic variation, and we have the following observations: 1) In the pre-trained SSL model, the acoustic dynamics result in UERs by up to 22.7\% (in rhythm), indicating the distinct alteration of derived representations. The pre-trained SSL model learns both linguistic and acoustic information given speech, and thus the units derived from speech with the same content can be indeterministic; however, 2) with the proposed bilateral perturbation (BiP), a distinct drop of UER (in energy) by up to 82.8\% could be witnessed, demonstrating the efficiency of BiP in producing deterministic representations referring to linguistic content.

\begin{table}[h]
    \centering
    \begin{tabular}{c|c|c}
    \toprule
    Acoustic   & Pretrained   & BiP-Tuned  \\
    \midrule
    Reference & 0.0     &   0.0   \\
    \midrule
    Rhythm $\hat{S_r}$  & 22.7   &  10.2     \\
    Pitch  $\hat{S_p}$  & 16.3   &  4.3      \\
    Energy $\hat{S_e}$  & 10.5   &  1.8     \\
    \bottomrule
    \end{tabular}
    \caption{We calculate UER between units derived from original and perturbed speeches respectively using the \textbf{pre-trained} and \textbf{fine-tuned} SSL model, which is calculated averaged over the dataset. It measures the ability of the SSL model to generate acoustic-agnostic representations referring to linguistic content.}
\label{table5}
\end{table}

\section{Evaluation on speech quality} \label{speech_quality}

Following the publicly-available implementation fairseq~\citep{ott2019fairseq}, we include the SNR as an evaluation matrix to measure the speech quality across the test set. We approximate the noise by subtracting the output of the enhancement model from the input-noisy speech and then compute the SNR between the two. Further, we conduct crowd-sourced human evaluations with MOS, rated from 1 to 5 and reported with 95\% confidence intervals (CI). For easy comparison, the results are compiled and presented in the following table:

\begin{table}[h]
\centering
\begin{tabular}{l|c|c}
\toprule
 Method & SNR ($\uparrow$) & MOS ($\uparrow$) \\
\midrule
Translation GT & / & 4.22$\pm$0.06\\
\midrule
DirectS2ST	 & 46.45 & 4.01$\pm$0.07\\
TextlessS2ST & 47.22  & 4.05$\pm$0.06\\
\midrule
TranSpeech & 46.56 & 4.03$\pm$0.06 \\
\bottomrule      
\end{tabular}
\caption{Speech quality (SNR($\uparrow$) and MOS($\uparrow$)) comparison with baseline systems. }
\label{appendix:speech_quality}
\end{table}

As illustrated in Table~\ref{appendix:speech_quality}, TranSpeech has achieved the SNR and MOS with scores of 46.56 and 4.03 competitive with the baseline systems. Since we apply the publicly-available pre-trained unit vocoder and leave it unchanged for unit-to-speech, we expect our model to exhibit high-quality speech generation as baseline models while achieving a significant improvement in translation accuracy.

\section{Information Enhancement} \label{perturbation}
\begin{figure}[ht]
    \centering
    \includegraphics[width=1.0\textwidth]{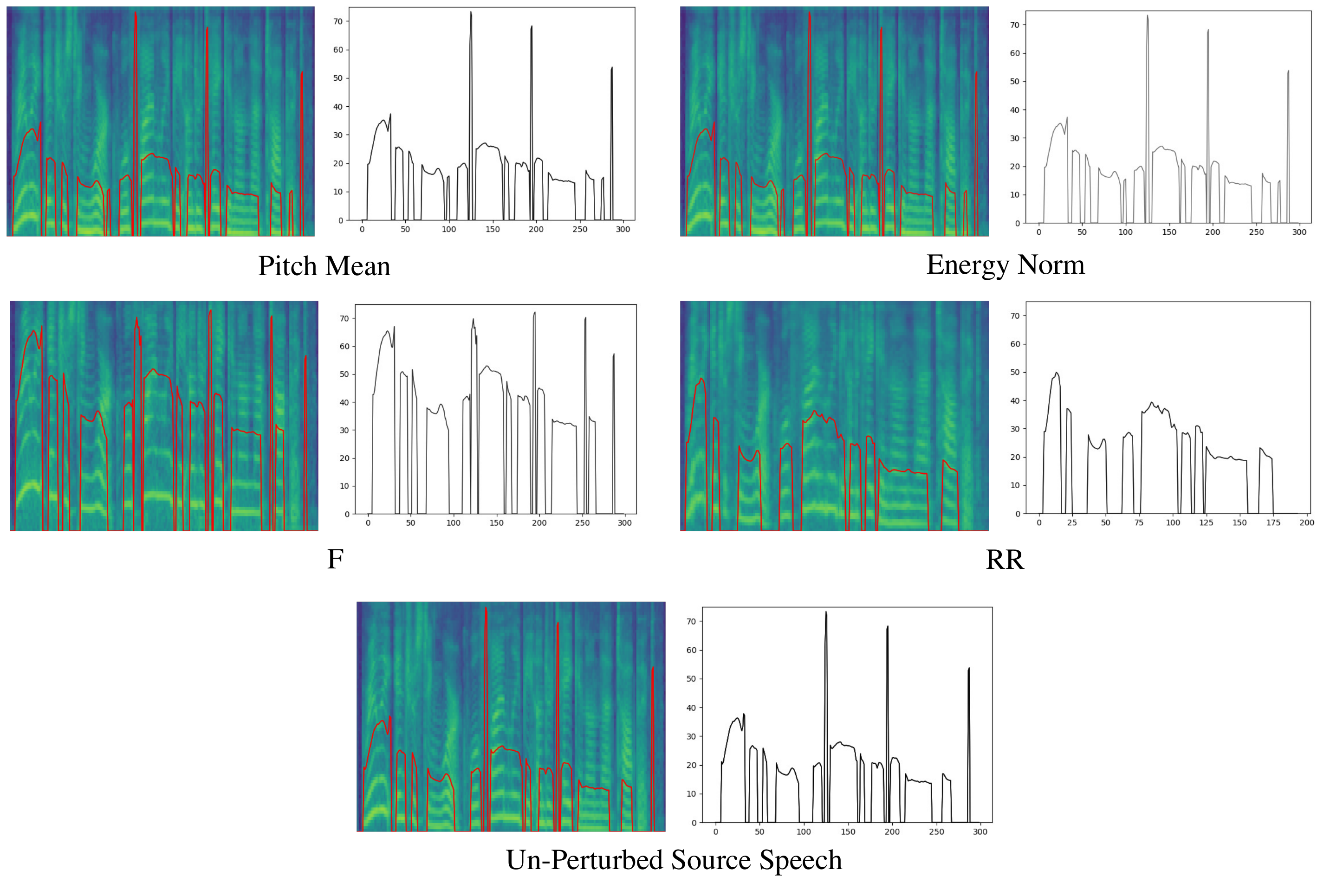}
    \caption{Spectrogram and pitch contours of the utterance with the single-perturbed acoustic condition, remaining the linguistic content ("really interesting work will finally be undertaken on that topic") unchanged. RR: random resampling. F: a chain function $F = fs(pr(peq(x)))$ for random pitch shifting. }
    \vspace{-2mm}
    \label{fig:perturbation}
\end{figure}

We apply the following functions~\citep{qian2020unsupervised,choi2021neural} on acoustic features (e.g., rhythm, pitch, and energy) to create acoustic-perturbed speech samples $\hat{S}$, while the linguistic content remains unchanged, including 1) formant shifting $fs$, 2) pitch randomization $pr$, 3) random frequency shaping using a parametric equalizer $peq$, and 4) random resampling $RR$. As shown in Figure~\ref{fig:perturbation}, we further illustrate the mel-spectrogram of the single-perturbed utterance in bilateral perturbation.
\begin{itemize} [leftmargin=*]
    \item For $fs$, a formant shifting ratio is sampled uniformly from $\mathrm{Unif}(1,1.4)$. After sampling the ratio, we again randomly decided whether to take the reciprocal of the sampled ratio or not. 
    \item In $pr$, a pitch shift ratio and pitch range ratio are sampled uniformly from $\mathrm{Unif}(1,2)$ and $\mathrm{Unif}(1,1.5)$, respectively. Again, we randomly decide whether to take the reciprocal of the sampled ratios or not. For more details for formant shifting and pitch randomization, please refer to Parselmouth~\url{https://github.com/YannickJadoul/Parselmouth}.
    \item $peq$ represents a serial composition of low-shelving, peaking, and high-shelving filters. We use one low-shelving HLS, one high-shelving HHS, and eight peaking filters HPeak.
    \item $RR$ denotes a random resampling to modify the rhythm. The input signal is divided into segments, whose length is randomly uniformly drawn from 19 frames to 32 frames~\citep{polyak2019attention}. Each segment is resampled using linear interpolation with a resampling factor randomly drawn from 0.5 to 1.5.
\end{itemize}

\section{Visualization of acoustic-perturbed speech samples} \label{vis_per}

\begin{figure}[h]
    \centering
    \small
    \includegraphics[width=1.0\textwidth,trim={1.0cm 0cm 1.0cm 0cm}]{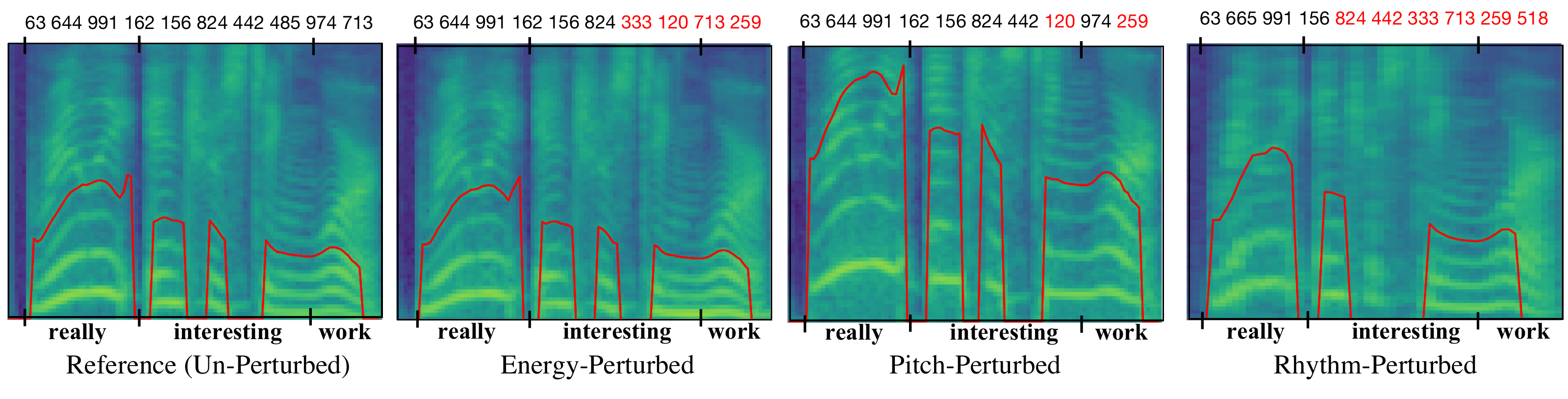}
    \caption{Spectrogram and pitch contours of speech sample with the perturbed acoustic condition, remaining the linguistic content ("really interesting work.") unchanged. The altered units are printed in red upside the spectrogram.} 
    \label{fig:info_perturbation}
\end{figure}

% \section{Potential Negative Societal Impacts} \label{impact}

% TranSpeech lowers the requirements for textless speech-to-speech translation, which may cause unemployment for people with related occupations such as interpretor and translator. In addition, there is the potential for harm from non-consensual voice cloning or the generation of fake media and the voices of the speakers in the recordings might be overused than they expect.

% \section{Reproducibility Statement}
% We will release our code in the future. The TranSpeech model that we build upon is publicly available through the fairseq code repository~\citep{ott2019fairseq}. To aid reproducibility, we have included a schematic overview of the algorithm in Figure~\ref{fig:norm}, and hyperparameters in Table~\ref{tab:hyperparameters}.

\end{document}